\documentclass[lettersize,journal]{IEEEtran}
\usepackage{amsmath,amsfonts,amssymb}
\usepackage{algorithmic}
\usepackage{algorithm}
\usepackage{array}
\usepackage[caption=false,font=normalsize,labelfont=sf,textfont=sf]{subfig}
\usepackage{textcomp}
\usepackage{stfloats}
\usepackage{url}
\usepackage{verbatim}
\usepackage{graphicx}
\usepackage{cite}
\usepackage{epsfig}  
\usepackage{makecell,multirow,diagbox}
\usepackage {hyperref}
\usepackage{booktabs}
\begin{document}

\title{\huge Unsupervised Learning Optical Flow in Multi-frame Dynamic Environment Using Temporal Dynamic Modeling }

\markboth{preprint to Arxiv.org ,~Vol.~XX, No.~XX, xxx~xxxx}%
{Shell \MakeLowercase{\textit{et al.}}: A Sample Article Using IEEEtran.cls for IEEE Journals}

\author{Zitang Sun, Shin'ya Nishida, and  Zhengbo Luo}



\maketitle

\begin{abstract}
For visual estimation of optical flow, a crucial function for many vision tasks, unsupervised learning,  using the supervision of view synthesis, has emerged as a promising alternative to supervised methods, since ground-truth flow is not readily available in many cases. However, unsupervised learning is likely to be unstable when pixel tracking is lost due to occlusion and motion blur, or the pixel matching is impaired due to variation in image content and spatial structure over time. In natural environments, dynamic occlusion or object variation is a relatively slow temporal process spanning several frames. We, therefore, explore the optical flow estimation from multiple-frame sequences of dynamic scenes, whereas most of the existing unsupervised approaches are based on temporal static models. We handle the unsupervised optical flow estimation with a temporal dynamic model by introducing a spatial-temporal dual recurrent block based on the predictive coding structure, which feeds the previous high-level motion prior to the current optical flow estimator. Assuming temporal smoothness of optical flow, we use motion priors of the adjacent frames to provide more reliable supervision of the occluded regions. To grasp the essence of challenging scenes, we simulate various scenarios across long sequences, including dynamic occlusion, content variation, and spatial variation, and adopt self-supervised distillation to make the model understand the object’s motion patterns in a prolonged dynamic environment. Experiments on KITTI 2012, KITTI 2015, Sintel Clean, and Sintel Final datasets demonstrate the effectiveness of our methods on unsupervised optical flow estimation. The proposal reaches state-of-the-art performance with advantages in memory overhead.

\end{abstract}


\section{Introduction}
\label{s1}
\IEEEPARstart{I}{n} computer vision,  motion perception is defined as optical flow estimation, a visual cue defined as the projection of the apparent motion of objects in a scene onto the image plane of a vision system or a visual sensor. As a fundamental vision task, optical flow estimation plays an essential role in various high-level vision tasks, such as video understanding\cite{jiang2018super}, behavior recognition\cite{simonyan2014two}, object tracking \cite{behl2017bounding}, etc.

\begin{figure}[!t]
	\centering
	\includegraphics[width=3.5in]{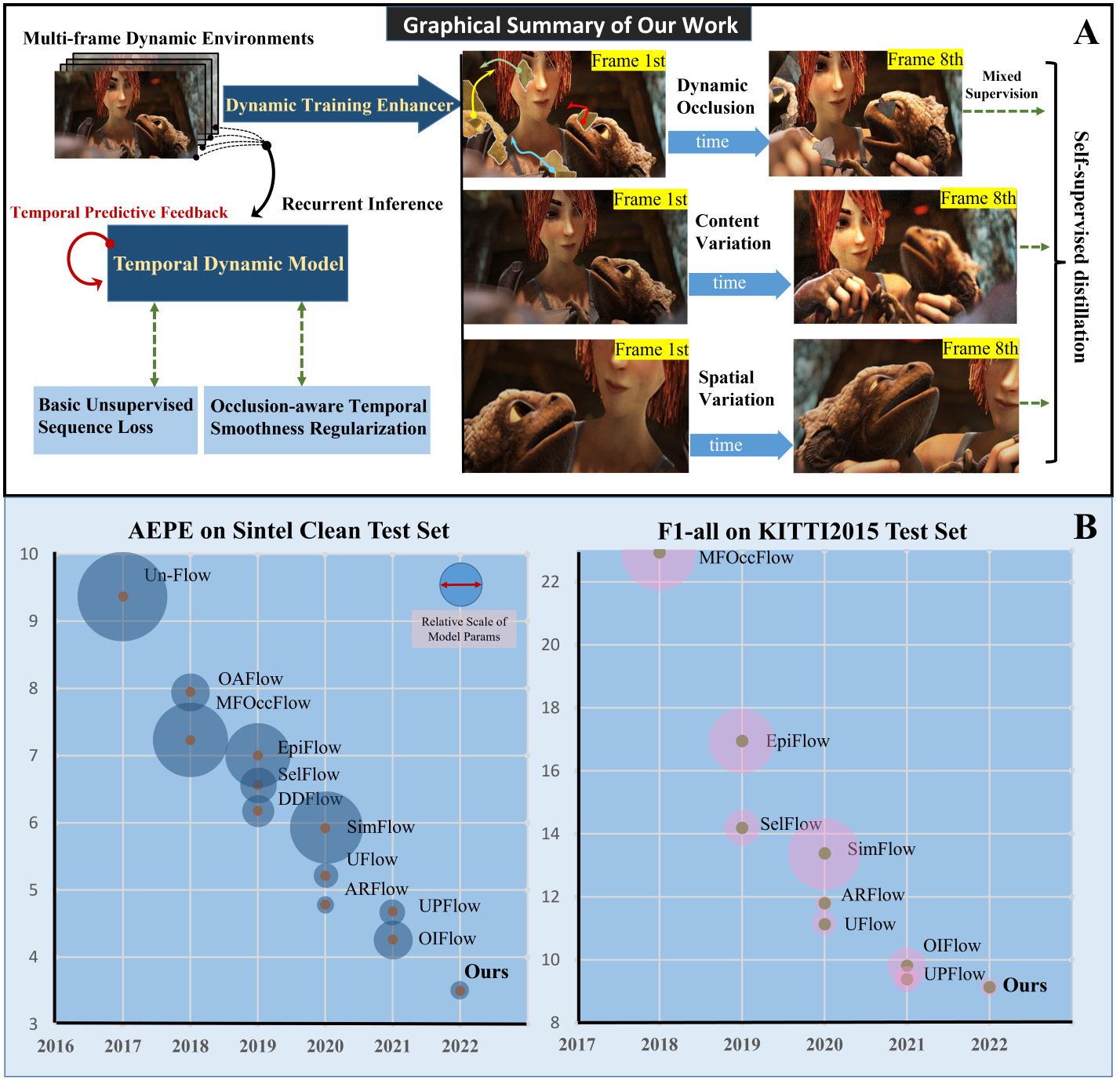}
	\caption{\textit{\textbf{Graphical summary of our work (A) and timeline of average end-point error (AEPE) and F1-all  in unsupervised optical flow estimation (B).}} \textbf{A}: The proposed temporal recurrent network understands optical flow estimation in dynamic environments. The relationships between different components are briefly described in the figure, where the Temporal Dynamic Model is introduced in subsection \ref{s4a}, the Occlusion-aware Temporal Smoothness Regularization is described in subsection \ref{s4b}, and the Dynamic Training Enhancer is proposed in section \ref{s5}  for  Dynamic Occlusion (\ref{s5a}), Content Variation and Spatial Variation (\ref{s5b}), and Self-supervised Distillation (\ref{s5c}), respectively; \textbf{B}: Marker size indicates network size, and oversized markers have been adjusted. Please refer to Table \ref{t1} and Table \ref{t2} for details.}
	\label{f0}
	\vspace{-0.5cm} %
\end{figure}

Early optical flow estimation focused on minimizing elaborate energy equations \cite{brox2004high,revaud2015epicflow}, which often suffers various limitations, including slow inference speed, inability to generate dense optical flows and handle complicated situations, etc. The performance of optical flow techniques has recently seen dramatic improvements with the  introduction of deep learning. FlowNet\cite{flownet} pioneered convolutional neural networks to address dense optical flow estimation, and a variety of approaches based on DNN have sprung up subsequently. Currently, dense optical flow estimation is divided into two main categories: supervised learning and unsupervised learning. Because ground-truth labels for dense optical flow are extremely difficult to obtain for real image pairs, supervised optical flow techniques are basically trained using synthetic data. Although models trained on synthetic data often generalize well to natural images, there is an inherent mismatch between two data sources, which those approaches may struggle to overcome\cite{goodflowdata}. Redundant label rendering and unreal data make unsupervised learning optical flow estimation increasingly desirable.

Unsupervised learning frameworks are proposed to utilize the resources of unlabeled videos \cite{unsupervisedlearingflow},
the overall strategy behind which is adopting a photometric loss that measures the difference between the target image and the (inversely) warped source image based on the dense flow field predicted from the network. Yet there are many challenges with image warping-based unsupervised learning, including brightness variations, color variations, and motion blur in multi-frame dynamic environments. More importantly, since images are two-dimensional projections of  3D space, the spatial occlusions induced by different objects in motion also cause pixel information loss over frames, which are unable to be recovered by image warping. All aforementioned factors largely mislead the networks and thus degrade the performance.

Several works have devoted to solving the above problems, including designing more robust loss functions \cite{2020deepfeaturematching,censusloss}, generating occlusion masks to circumvent the occluded region's supervision\cite{wang2018occlusion}, tracking the occluded pixels in adjacent frames \cite{2018unsupervisedmultiframe}, and artificially adding random noise to simulate the occlusion scenes \cite{selflow}. Besides, there is a growing tendency to improve performance by integrating multiple strategies \cite{uflow,arflow,luo2021upflow}. 

We notice that most approaches are based on a pair of images to estimate optical flow. Although a few works utilized multiple frames\cite{2018unsupervisedmultiframe,arflow}, the structure is still temporally static feedforward CNNs per se. As humans perceive motion and understand the world from dynamic environments in an unsupervised manner, here we hold the view that neural networks for motion perception should also be trained in extended temporal dynamic environments.
Structurally, we partially refer to temporal predictive coding in human visual systems\cite{predictivecoding,prednet}, where higher-order neurons send feedback signals to control lower-order neurons. In continuous motion of objects, the loss of pixels information due to occlusion and blurring, which may not be a problem in supervised learning due to the presence of ground truth from God's view, is untraceable and tricky in unsupervised learning. The predictive coding structure based on temporal dynamics has at least the following potential advantages for alleviating the problem. First, Based on the motion temporal smoothness, temporal dynamic networks can propagate the previous motion prior through hidden states to provide a more reliable reference for untraceable pixels. Second, the motion estimation of the occluded object is practically equivalent to the prediction of the location of the occluded object in the latter frame, which has a strong connection with predictive coding mechanism as one of the motivations for introducing temporal dynamics.

 Accordingly, this work constructs an CNN-RNN based model that understands optical flow estimation in dynamic environments. Recent supervised learning work RAFT\cite{raft} demonstrates the strength of recurrent blocks in optical flow estimation. Instead of recurrent refinement of a single frame, we introduce both spatial and temporal recurrences into the model, constructing a spatial-temporal dual recurrent block with predictive coding arrangement. The final resulting network has a small number of parameters yet stably converges in the sequence of arbitrary length and demonstrates higher performance in ablation study compared to other models. 

We trained the model in the temporal causal sequence. To make the model correctly grasp the object occlusion, brightness variation, color variation, and various content blurring caused by the movement in the dynamic environment, three training enhancers based on self-supervised learning are proposed, including \textbf{Dynamic Occlusion Enhancer} (DOE), \textbf{Content Variation Enhancer} (CVE), and \textbf{ Spatial Variation Enhancer} (SVE). In which DOE extracts sub-object blocks from the original data distribution and simulates the random natural motion of objects in multi-frame images, allowing the model to understand object occlusion patterns over prolonged temporal environments. Furthermore, we propose a mixed supervision strategy by combining the unsupervised loss and self-supervised loss in the DOE, which simultaneously provides reliable supervision on the occluded region and the occlusion per se, facilitating a better generalization of the network to occlusion scenes. CVE simulates the variation of image content, such as brightness and chromaticity of the object across time. Also, it simulates motion blur, defocuses blur, etc., caused by high-speed moving, boosting the robustness of the model in dealing with such challenging scenes. The final SVE simulates the spatial variation of the object that is often caused by camera shake and environmental vibration in dynamic environments, including random rotation, scaling, translation, and other series of affine transformations.

To solve the occlusion issue, the majority of the works utilize occlusion masks to circumvent the supervision of the occluded region, which, nonetheless, results in inadequate supervision of the occluded region. Based on a multi-frame training environment, we propose a novel temporal smoothness regularization to alleviate this problem by assuming approximately the constant velocity of identical objects within a short temporal window (due to object inertia), which provides more reliable motion prior to the occluded regions. 

Combined with the proposed multiple strategies, we eventually achieve the state-of-the-art performance on multiple datasets with a lightweight network configuration, as shown in Fig. \ref{f0}.

Our strategy is motivated by multiple recent works, including \cite{arflow,uflow,starflow,janai2018unsupervised}. The contributions in this study are listed as follows:

\begin{itemize}
\item We introduce temporal dynamic modeling to the unsupervised learning-based optical flow estimation task and design an efficient CNN-RNN based on a predictive coding structure that can process the sequence of arbitrary length recursively, achieving state-of-the-art results on several leading datasets, as shown in Fig. \ref{f0}.

\item We build three self-supervised training enhancers based on multi-frame dynamic environments, which substantially improve the generalization performance in various challenging environments by reducing 20\% errors in the Sintel dataset. The method is also flexible and compatible with other multi-frame methods to improve model performance.

\item We propose two strategies for solving the occlusion problem, including a mixed supervision strategy and temporal smoothing regularization,  alleviating the problem of inadequate supervision for occluded regions.

\end{itemize}
The remaining part of this work is organized as follows: In Section \ref{s2}, recent trends of unsupervised learning-based optical flow estimation approaches are briefly reviewed. The preliminaries of unsupervised learning based approach and our notion convention are introduced in Section \ref{s3}. The proposed network structure is introduced in Section \ref{s4}. 
Section \ref{s5} presents the proposed three types of self-supervised learning-based training enhancers. Experiments and ablation studies are given in Section \ref{s6}. Finally, we conclude this work in Section \ref{s7}.

\section{Related work}
\label{s2}
\textbf{Supervised methods} require annotated flow ground truth to train the network. FlowNet \cite{flownet} is the first work to propose learning optical flow estimation by training a fully convolutional network on the synthetic dataset FlyingChairs. Then, FlowNet2 \cite{flownet2} proposed stacking multiple networks as an iterative improvement. SpyNet \cite{spynet} built a spatial pyramid network to estimate optical flow in a coarse-to-fine manner to cover challenging scenes with large displacements. PWC-Net \cite{pwcnet}, and LiteFlowNet \cite{hui2018liteflownet} proposed to warp features and calculate the cost volume of each pyramid layer by building efficient and lightweight networks. IRR-PWC \cite{hur2019iterative} proposes to design pyramid networks by iterative residual refinement schemes. Recently, RAFT \cite{raft} proposed to estimate flow fields by 4D correlation volume and recurrent network, yielding an excellent performance. 

\textbf{Unsupervised approaches} circumvent the need for labels by optimizing photometric consistency with some regularization.
Yu et al.\cite{unsupervisedlearingflow} first introduced a method for learning optical flow with brightness constancy and motion smoothness, which is similar to the energy minimization in conventional methods. Further researches improve the accuracy through occlusion reasoning \cite{wang2018occlusion,censusloss}, multi-frame extension \cite{janai2018unsupervised,guan2019unsupervised}, epipolar constraint \cite{epipolarflow}, 3D geometrical constraints with monocular depth \cite{dfnet,depthflowcamera} and stereo depth \cite{wang2019unos,liu2019unsupervised}. Although these methods have been complicated, there is still a large gap with state-of-the-art supervised methods. Based on self-supervised learning, recent works improve the performance by learning the flow of occluded pixels in a knowledge distillation manner\cite{liu2019ddflow,selflow,arflow}. There is a growing tendency to improve performance by integrating multiple strategies \cite{uflow,luo2021upflow}.

\textbf{The treatment of occlusion} has been a fundamental problem of unsupervised learning. The most common approach is to use an occlusion mask to ignore the occluded regions' photometric loss \cite{wang2018occlusion}, which, nevertheless, results in the occluded regions never getting effective supervision. In self-supervised learning, some methods add random noise to the latter frame as a simulation of the occlusion scene\cite{liu2019ddflow,selflow,arflow}. However, occlusion is a long temporal process that varies in the interaction among different moving objects. Sudden masking by random noise of the above approaches does not follow the smooth motion of objects in natural scenes that obey the inertia criterion. Also, images contaminated by a large part of random noise can be regarded as an outlier outside the distribution of the original dataset, which has potential factors to degrade the network's performance on the original dataset.
In this work, we proposed a new strategy that extracts sub-objects from original data distribution and simulates a smooth dynamic occlusion process in self-supervised training. With a well-designed mixed loss function, the full supervision of flow both on occluded regions and occluder per se is realized.

\textbf{Temporal dynamic models} based on recurrent neural networks are not dominant approaches in optical flow estimation.
A few methods have tried to model multi-frame optical flow estimation using RNN and LSTM\cite{starflow,neoral2018continual}, which employ conventional supervised training strategies and can only be applied to a limited number of synthetic sequences.
 There are also several approaches in unsupervised learning that utilize multi-frame to estimate optical flow\cite{2018unsupervisedmultiframe,guan2019unsupervised,arflow,selflow}. However, essentially they are still temporally static networks and cannot handle real-time causal sequences with arbitrary lengths. In this work, we introduce the predictive coding structure and construct a spatial-temporal dual recurrent network that unsupervisedly learns optical flow in a prolonged temporal environment. Moreover, we combine existing self-supervised methods and design a novel dynamic training environment based on confidence propagation, driving the network more capable of handling complex dynamic scenarios.

\section{Preliminaries and Notation}
\label{s3}
In this section, we first briefly introduce the general framework of the unsupervised optical flow used in our approach and define the basic notation used in subsequent sections.
Let casual RGB image sequence $\mathcal{I}=\{I_0,I_1,...,I_{t-1},I_{t}\}$, our target is to train a model that estimate current optical flow $F_{t-1\rightarrow t} \in \mathbb{R}^{H \times W \times 2}$ from $\mathcal{I}$, i.e. , $F_{t-1\rightarrow t} = f (\mathcal{I};\Theta)$, where $\Theta$ is a set trainable parameters of the model.

We follow the general warping-based strategy as our basic unsupervised learning approach, by which the network can be trained implicitly with view synthesis. Specifically, each image $I_t$ could be reconstruct by its next frame $I_{t+1}$ via inverse warping operation, i.e., $\hat{I}_t = I_{t+1}(\mathbf{p}+F_{t\rightarrow t+1}(\mathbf{p}))$, where $\mathbf{p}\in \mathbb{R}^{H\times W}$ is the spatial position across whole images. Utilizing bilinear interploation, the method can be realized in a differentiable way at sub-pixel level. Then one only needs to define the similarity evaluation function $\rho(\cdot)$ between the reconstructed image $\hat{I}_t$ and the original image${I}_t$, i.e. $\mathcal{L}_{\mathrm{ph}} \sim \sum_{\mathbf{p}} \rho(\hat{I}(\Theta), I)$, so as to implicitly supervise the generated optical flow. The basic similarity evaluation loss is L1, or Charbonnier loss\cite{sun2010secrets}, which simply considers the pixel-wise similarity between a pair of images. The more robust loss that focuses on structural similarities, like SSIM loss and Census loss\cite{censusloss}, are wildly used in the community to overcome the variations in color and brightness across frames.

Due to the aperture problem and the ambiguity of local appearance, supervision solely based on the photometric loss does not sufficiently constrain the problem for somewhere textureless or with repetitive patterns. One of the most common ways to reduce ambiguity is named edge-aware first and second-order smooth regularization\cite{tomasi1998bilateral}. Given frame $I_{t-1} $ with flow field  $F_{t-1\rightarrow t}$, the  $k$-order edge-aware loss $\mathcal{L}_{s m(k)}$ could be defined as:

\begin{figure*}[!t]
	\centering
	\vspace{-0.5cm} %
	\includegraphics[width=7in]{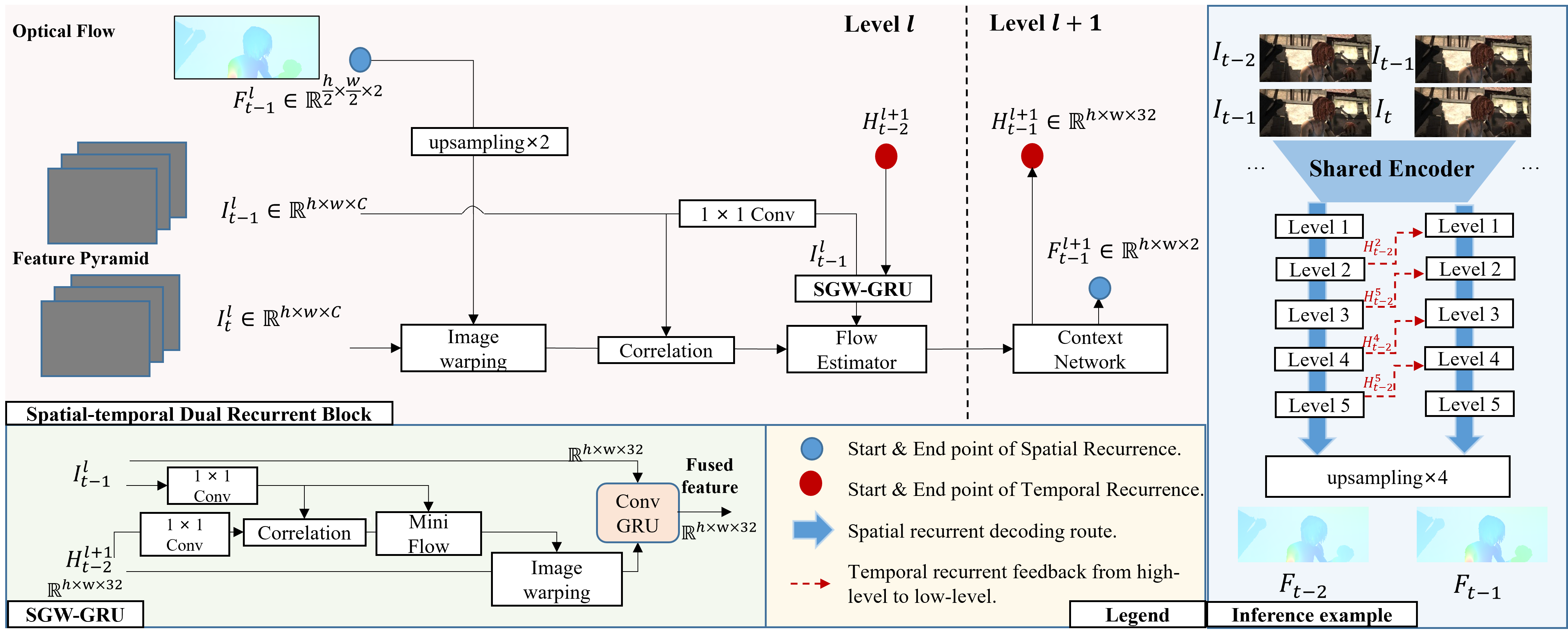}
	\vspace{-0.2cm} %
	\caption{\textit{\textbf{The inference architecture of the proposed spatial-temporal dual recurrent block.}} The spatial recurrence is modified from the basic flow inference pipeline of PWCNet\cite{pwcnet}, including a cascade of image warping, correlation volume, flow estimator, and context network, which decodes the flow field from small to large scales. By designing a self-guided warping-based GRU (SGW-GRU) block, we introduce the temporal recurrence based on a predictive coding structure that transmits feedback from deep to shallow layers. The right subfigure demonstrates that the network can infer temporal causal sequences with arbitrary length. Please view subsection \ref{s4a} for details.}
	\vspace{-0.5cm} %
	\label{f1}
\end{figure*}
\begin{equation}
	\begin{aligned}
	&\mathcal{L}_{s m(k)}\sim\sum_\mathbf{p}\left(w_\mathbf{p}^x\cdot\left|\frac{\partial^{k} {F}_{t-1\rightarrow t}}{\partial x^{k}}\right|_\mathbf{p}+w_p^y\left|\frac{\partial^{k} {F}_{t-1\rightarrow t}}{\partial y^{k}}\right|_\mathbf{p}\right),\\
	&w_\mathbf{p}^x =  \exp \left(-{\lambda} \sum_{c}\left|\frac{\partial I_{t-1}}{\partial x}\right|_\mathbf{p}\right),\\
	&w_\mathbf{p}^y =	\exp \left(-\lambda \sum_{c}\left|\frac{\partial I_{t-1}}{\partial y}\right|_\mathbf{p}\right).\\
\end{aligned}
\label{e1}
\end{equation}
Where $w_\mathbf{p}^x$, $w_\mathbf{p}^y$ is the attenuation weight determined by the smoothness of the first-order image gradient, which intuitively forces the flow field in areas with similar pixels to be smoother. Following previous work\cite{arflow,uflow}, we use first-order smooth loss on Sintel dataset, second-order smooth loss on KITTI dataset.

Motion occlusion is a challenging problem in unsupervised learning, which results in pixel-wised information loss from $I_t$ to $I_{t-1}$.
The most common method is to circumvent the loss calculation for the occluded region by inferring the binary occlusion mask $O_{t-1\rightarrow t} \in \{0,1\}^{H\times W}$ from $I_{t-1}$ to $I_t$, where $O_{t-1\rightarrow t}(\mathbf{p}) =0 $ means the locations $\mathbf{p}$ from $I_{t-1}$ are occluded in $I_{t}$ and vice versa. Following the previous method\cite{uflow,arflow}, a forward-backward check\cite{fbcheck} is used in our basic approach to estimate the occlusion mask.

We simplify the notion in the following section. Given two adjacent time step $\{a,b \mid b>a\}$, one denotes forward flow $F_{a\rightarrow b}$ as $F_{a}$, and backward flow $F_{b\rightarrow a}$ as $F^{-1}_{b}$; $\mathbf{p} \in \mathbb{R}^{H\times W} $ is denoted as spatial position set of a specific image; $\mathcal{W}_{a}(\cdot) :I_{b}\longmapsto \hat{I}_{a} $ is denoted as warping operation from $I_b$ via $F_{a \rightarrow b }$; $\mathbf{E}_\mathbf{p}(\cdot)$ represents the statistical expectations along the $\mathbf{p}$.

\section{Basic approach }
\label{s4}
In this section, we introduce the proposed spatial-temporal dual recurrent network and our basic unsupervised training approach based on temporal casual sequence. Also, a novel occlusion-aware temporal smoothness regularization is introduced in subsection \ref{s4b}. The effectiveness of several proposal would be demonstrated in ablation study part. 

\subsection{Network structure}
\label{s4a}
We need to construct a dynamic RNN model that can handle arbitrary temporal series $\mathcal{I}=\{I_0,I_1,...,I_{t-1},I_{t}\}$, which takes the previous frame $I_{t-1}$, the current frame $I_{t} $, as well as the previous hidden state $H_{t-2}$ as input, and outputs the current optical flow  $F_{t-1}$, and the hidden state $H_{t-1}$ served for the next time step.

In order to introduce temporal recurrence, we design the convolutional GRU block with a self-guided warping operation, namely the SGW-GRU block.
Given a pair of images, a five-level feature map pyramid  $\mathcal{P} = \{I^5,...,I^1\}$ are generated by a concise feedforward CNN for each image respectively, with size gradually reducing from  $\frac{H}{4}\times\frac{W}{4} $ to $\frac{H}{64}\times\frac{W}{64} $.  The SGW-GRU block directly operates on feature pyramids from low-scale to large-scale in a spatially recurrent manner. Also, it operates across image sequences in a recurrent temporal manner. For feature pyramid at level $l$, the spatial recurrent process basically follows the four stages of PWCNet\cite{pwcnet}: warping $\rightarrow $ correlation volume $\rightarrow $ flow estimator $\rightarrow $ context network, as shown in Fig. \ref{f1}. 
The main difference is that an extra convolutional block is plugged into the context network for generating the new hidden state $H_{t-1}$. The hidden state is characterized by 32-channel feature maps at different scales, which implicitly contain higher-order motion prior and contextual information from previous moments, served as input for the next time stage by feeding into the SGW-GRU block. Due to the object's motion, there is constantly a difference in spatial location between the features of two adjacent time steps. Accordingly, we design a self-guided warping (SGW) block that enables the model to adjust the spatial location of hidden feature maps by itself. In SGW blocks, the correlation volume between the previously hidden feature and the current feature is first calculated, based on which a mini flow estimator is subsequently employed to estimate a flow field, guiding the warping operation to adjust the spatial arrangement of $H_{t-2}^{l+1}$. Finally, a convolutional GRU module fuses the hidden state  $H_{t-2}^{l+1}$ with the current features $I_{t-1}^l$ and outputs the fused feature, as shown in the lower-left corner of  Fig. \ref{f1}.
Conv-GRU cell basically follows the principle of the original Gate Recurrent Unit\cite{gru} but replaces the operation with fully convolutional approaches. Specifically, it takes previous high-level hidden state $H_{t-2}^{l+1}$ and current feature $I_{t-1}$ as input and generates fused feature $I_{new}$ served as input for flow estimator, the process of which could be formalized as:
\begin{equation}
	\begin{aligned}
		&z=\sigma\left(\operatorname{Conv}_{3 \times 3}\left(\operatorname{ConCat}\left[H_{t-2}^{l+1}, I_{t-1}\right]\right)\right) \\
		&r=\sigma\left(\operatorname{Conv}_{3 \times 3}\left(\operatorname{ConCat}\left[H_{t-2}^{l+1}, I_{t-1}\right]\right)\right) \\
		&q=\operatorname{Tanh}\left( \operatorname{Conv}_{3 \times 3}(\operatorname{ConCat}[r\odot I_{t-1} , H_{t-2}^{l+1}])\right)\\
		&I_{new}=\left(1-z\right) \odot I_{t-1}+z \odot q.
	\end{aligned}
\end{equation}
Where $\sigma(\cdot) $ denotes sigmoid function; $\operatorname{ConCat}(\cdot)$ denotes concatenation of feature maps; and $\odot$ means element-wise production.
The other difference is that Conv-GRU cells do not generate new hidden states, which are instead realized by context networks subsequently.

In the structure of temporal recurrence, the feedback signal from the end of the SGW-GRU block is transferred to the head in shallow layers at the next time state to control its coding behavior, so the neurons at the current time state would predictively consider future motion patterns, as shown in the right side of Fig. \ref{f1}, the mechanism behind which is similar to the temporal predictive coding structure in the human visual system \cite{predictivecoding,prednet},  where higher-order neurons send feedback signals to lower-order neurons and control their behavior. This mechanism is considered to implicitly propel neurons to learn motion perception in a dynamic environment\cite{2021learnmaterial}.

\begin{figure*}[!t]
	\centering
    \vspace{-0.4cm} %
	\includegraphics[width=7in]{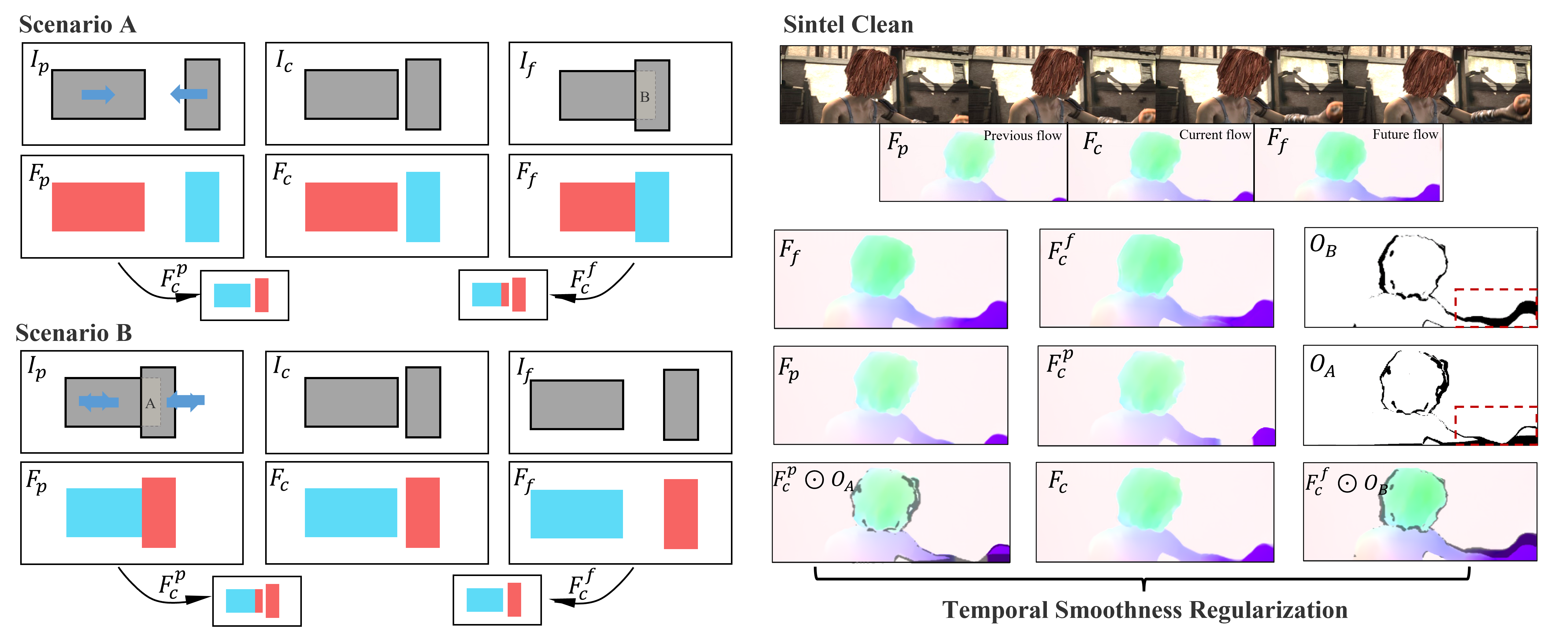}
	\caption{\textbf{\textit{Illustration of Occlusion-aware Temporal Smoothness Regularization.}} 
		\textbf{Scenario A}: pixels in the current frame $I_c$ become invisible in the next frame $I_f$; \textbf{Scenario B}: pixels visible in the current frame $I_c$ are invisible in the previous frame $I_p$. Both Scenario A and B cause errors in spatial alignment, which should be considered in temporal smoothness regularization. The right side shows the actual Sintel clean scene. We introduce a temporal smoothing constraint on the current flow field $F_c$ by tracking the motion prior from previous $F_p$ and future frames $F_f$ under the assumption of constant velocity within short time windows. Due to the smoothness of the motion, the object's forward and backward occlusions are usually complementary, which means that it is possible to track the motion prior as reference alternatively in $F_p$ or $F_f$, as shown in regions marked by the red dashed box.
		For more details, please access subsection \ref{s4b}}
		\vspace{-0.4cm} %
	\label{f2}
\end{figure*}

The integration of the above proposals leads to a lightweight network with only $2.50$ M parameters, which can handle causal sequences of arbitrary length and exhibits fair effectiveness in ablation study.

\subsection{ Occlusion-aware Temporal Smoothness regularization }
\label{s4b}
As mentioned in Section \ref{s1}, simply masking the occluded region causes the inadequate supervision for the occluded region, as where only supervised by spatial smoothing loss that simply evaluates the low-level RGB similarity. We propose a multi-frame-based temporal smoothness regularization to alleviate this problem as an alternative supervision method.

First we define four adjacent frames $\{ I_{k-1}, I_{k}, I_{k+1}, I_{k+2} \}$ and three motion flow field $\{F_{p}, F_{c}, F_{f} \}$ as forward optical flow of $I_{k-1}\rightarrow I_{k}$, $I_{k}\rightarrow I_{k+1}$ and, $I_{k+1}\rightarrow I_{k+2}$, respectively. Due to the limitation of physical laws, the motion trajectory $S(t)$ of an object is always a differential curve with smoothness, and the video sequence can be regarded as a sampling of its locational state. With a sufficient short sampling interval $\Delta t$, one can pick a small temporal window, and make a Taylor first-order approximation of the $S(t)$, i.e., $S(t) = S(t_0) +  \Delta t \cdot \left.\frac{\mathrm{d} S}{\mathrm{~d} t}\right|_{t=t_0}$, or, $S(t) = S(t_0) +  t \cdot V(t_0)$, where $V$ could be regard as velocity vector or so-called optical flow within the defined temporal window.

As a result, a warping operation based on the flow field of the current frame $F_c$ can be used to spatially align the flow field of the previous frame $F_p$ and the future frame $F_f$, which can be represented by the following equation:

\begin{equation}
	\begin{aligned}
	 &\hat{F}_c^{f} = F_{f}(p+F_{c}(p))\\
	&\hat{F}_c^{p} = F_{p}(p+F^{-1}_{p}(p)).\\
	\end{aligned}
\end{equation}

Where $\hat{F}_c^{f}$ represents reconstructed  $F_c$ from $F_f$, and $F^{-1}_{p}$ denotes backward flow of  $I_{k}\rightarrow I_{k-1}$. 
Under the aforementioned assumption of velocity constancy, the three adjacent optical flow fields can be approximately aligned in space, and their derivation along time gives the basic temporal smoothness loss function. However, an important issue is that warping is not a reversible operation, and there is an occlusion problem in either the forward or backward optical flow, as shown on the left side of Fig. \ref{f2}.
Typically, there are two problem scenarios, the first in which pixels in the current frame become invisible in the next frame (scene A), and the second in which pixels visible in the current frame are invisible in the previous frame (scene B). Both of these occlusion regions cause errors in spatial alignment.
For simplicity, the $\Omega_A$, $\Omega_B$ are denoted as two location sets that pixels of $F_c $ are invisible in the $F_p$ and $F_f$, respectively. Accordingly, we generate two occlusion masks  $O_A, O_B $ via forward-backward check that follows: 

\begin{equation}
	O(\mathbf{p})= \begin{cases}
		1, & \mathbf{p}\notin  \Omega\\
		0 & \mathbf{p}\in  \Omega
	\end{cases} 
\end{equation}

Temporal smoothness depends on the computation of the differential along time, which in the discrete case is equivalent to the computation of the first-order difference across frames. Here we simplify it by directly using Charbonnier loss to evaluate the difference between frames, thus ensuring the smoothness between $F_c$ and the two adjacent frames. Besides, the  displacement size of the $F_c$ is utilized as the decay coefficient for temporal smoothing, i.e., locations with more violent motion have smaller smoothing constraints, and the final temporal smoothing regularization of current flow $F_c$ is defined as:
 
 \begin{equation}
 \begin{aligned}
 \ell_{tsm} \sim&\frac{\sum_{\mathbf{p}}  \left[\mathbf{C}\left({\mathcal{S}}(\hat{F}_c^{p}), F_c\right)\odot O_A \bigg/
 	 |F_c| \right]}{\sum_{\mathbf{p}}O_A(\mathbf{p})} +\\
& \frac{\sum_{\mathbf{p}}  \left[\mathbf{C}\left({\mathcal{S}}(\hat{F}_c^{f}), F_c\right)\odot O_B\bigg/
	|F_c|\right]}{\sum_{\mathbf{p}}O_B(\mathbf{p})}.
 \end{aligned} 
 \end{equation}
Where $\mathcal{S} (\cdot)$ means stop the gradient from computational graph, $\mathbf{C} (\cdot)$ is Charbonnier loss.
As for locations $\mathbf{p} \in \{\Omega_A \cap \Omega_B\} $, we do not add any temporal smoothness regularization since no reliable flow priors exist in adjacent frames; As for positions $\{ \mathbf{p} | \mathbf{p} \in \Omega_A, \mathbf{p} \notin \Omega_B \}$, we only add temporal smoothness from the previous flow $F_p$; Similarly, temporal smoothness from the future flow $F_f$ only available for positions $\{ \mathbf{p} | \mathbf{p} \in \Omega_B, \mathbf{p} \notin \Omega_A \}$;  Eventually temporal smoothness from both $F_p$ and $F_f$ would be added to positions $\{ \mathbf{p} | \mathbf{p} \notin \Omega_A \cup \Omega_B \}$, and the optimized result of the last situation is that the $I_c$ approximate the value of linear interpolation between $I_p$ and $I_f$.

Our design includes a priori that due to the temporal smoothness of motion pattern, $\Omega_A$ and $\Omega_B$ tend to have only a few intersections, i.e., objects occluded in the next frame are often visible in the previous frame. Therefore, motion information occluded in forwarding propagation can be alternatively supplied via the past flow. 
As in the lower right side of Fig. \ref{f2}, the background area occluded by the arm in the forward inference is visible in the backward inference, so the motion information of its previous frame can be tracked using the backward flow as a reference. This temporal smoothness constraint provides more reliable supervision of the occluded region compared to spatial smoothness.

\subsection{ Training in Temporal Sequence }
\label{sc12}
In training, we feed $N$ consecutive frames into the network sequentially per iteration and simultaneously supervise the generated $N-1$ optical flows.  The combined strategies from Section \ref{s3} are adopted to construct the loss function, i.e., warping-based image photometric similarity, spatial smoothing loss, and occlusion masking.
It is worth noting that since our method is a temporal-recursive network, it can only infer causal sequences and generate optical flow in one direction. Therefore, in training, we first forward infer the forward optical flow for the whole sequence and then reverse the image sequence to infer backward optical flow sequences. Finally, the forward-backward check is practiced to obtain the occlusion mask sequence $\{O_t\}_{t=1}^{N-1}$.

Specifically, at the end of the sequence, we update the weights so as to decrease:

\begin{equation}
	\mathcal{L}=\frac{1}{N-1} \sum_{t=1}^{N-1} \mathcal{L}_{t}
\end{equation}
where 	$\mathcal{L}_t $ is combination of a multi-scale loss for image pair $(I_t, I_{t+1})$ and the temporal smoothness loss, which could is represented as:

\begin{equation}
	\label{e7}
	\begin{aligned}
	\mathcal{L}_t=&\lambda_{1}\ell_{sm}(F_t)+\lambda_{2}\ell_{tsm}(F_{t-1},F_t,F_{t+1})+\\
	&\ell_{warp}(I_t,I_{t+1},F_t,O_t).
	\end{aligned}	
\end{equation}
The $ \lambda_{1}$ and $\lambda_{2}$ are used to balance the loss weight as hyper-parameters. Please refer to  Section \ref{s6b} for more detailed configuration of loss function.

\section{Dynamic Training Enhancer}
\label{s5}
As proposed by DDFlow\cite{ddflow}, self-supervised learning-based distillation is an effective means to improve unsupervised optical flow estimation, which has been adopted in a large amount of work\cite{selflow,arflow,uflow,luo2021upflow}.
Generally, it depends on one teacher model $f_t(\cdot) $ that processes the original pair of samples $\mathcal{I} = \{I_a, I_b\}$ to generate optical flow $F_a^*$ as a pseudo label for training a student model $f_s(\cdot) $. 
 Before $\mathcal{I}$ is fed to the $f_s(\cdot)$, a series of specific image transformations $\mathcal{T}_I(\cdot): \mathcal{I} \longmapsto \widetilde{\mathcal{I}}$ can be used to increase the scene's diversity as a kind of data augmentation to generate optical flow $ F_a$ with lower confidence, which will be self-supervised by the pseudo label from $f_t(\cdot)$. To  ensure the spatial consistency, the pseudo label is required to be transformed consistently by  $\mathcal{T}_F(\cdot) : F^* \longmapsto \widetilde{F^*}$. After that,  the self-supervision based on  pseudo label can be realized  by considering Charbonnier similarity between $\mathcal{T}_F(F_a^*)$ and $F_a$, as:
 \begin{equation}
 	\ell_{self} \sim \mathbf {E}_{\mathbf{p}}\left(\left|\mathcal{S}\left({F}_{a}(\mathbf{p})\right)-\widetilde{F_{a}^{*}}(\mathbf{p})\right|+\epsilon\right)^{q}
 \end{equation}

In this work, we extend this strategy to multi-frame sequences and simulate multiple dynamic variation scenarios in nature, thus making the network understand the occlusion and variation in a dynamic environment. Specifically, we design three training enhancers, namely  {\bfseries Dynamic occlusion enhancer} (DOE), {\bfseries Spatial variation enhancer} (SVE), and {\bfseries Content variation enhancer} (CVE), which would be described as follows.

\subsection{Dynamic Occlusion Enhancer}

\label{s5a}
We integrate the occlusion hallucination into our framework, namely occlusion enhancer. Specifically, there are two steps: $(i)$ Random crop. Actually, the random crop belongs to spatial transformation, but it efficiently creates new occlusion in the boundary region. We crop the sequence of images as a preprocess of occlusion transformation. $(ii)$ Dynamic occlusion simulation.
Given a set of frame sequences, we divide the image into multiple sub-regions by using a super-pixel segmentation algorithm\cite{slic}.
Subsequently,   $n$ regions are randomly selected from the sub-regions as {\bfseries ``occluders"}. Instead of using random noise, we create natural textures for each occluder by extracting the texture from original sample batches and adding random noise while using a large kernel Gaussian filter to make the texture spatially correlated. Finally, we simulate the occlusion by randomly placing $n$ occluders into any spatial position of the first image.

\subsubsection{Simulation of Dynamic Motion Process }
Unlike previous work, we simulate the motion of the occluders dynamically in multi-frame sequences. Obviously, objects' motion always keeps smoothness in natural environments, and the object's motion at the next time step depends heavily on the motion state of the current moment. To better simulate random motions while keeping the above rule, we partially follow the Markov chain principle to simulate the process.

Specifically, we assume the moving state of an object as random process $\mathbf{S} (t)$, which can be decomposed two orthogonal vector $\mathbf{S}(t) = [U(t),V(t)]$. For any time step $t$, we introduce Markov properties by assuming its moving state only depends on the motion state of the previous time step $t-1$, i.e.:
\begin{equation}
	\begin{aligned}
		&\operatorname{Pr}\left[\mathbf{S}(t)=s_{t}\mid\mathbf{S}(t-1)=s_{t-1},\mathbf{S}(t-2) =s_{t-2}, \ldots\right] \\
		&\quad=\operatorname{Pr}[\mathbf{S}(t)=s_t \mid \mathbf{S}(t-1)=s_{t-1}]
	\end{aligned}
\end{equation}
The motion states $[U,V]$ at time $t$ are sampled from a two dimensional Gaussian distributions:
\begin{equation}
[U(t),V(t)] \sim \mathcal {N}(	\boldsymbol{\mu},\boldsymbol{\Sigma})
\end{equation}
where we let 
\begin{equation}
	\boldsymbol{\mu}=\left(\begin{array}{l}
		\mu_{U}(t) \\
		\mu_{V}(t)
	\end{array}\right), \quad \boldsymbol{\Sigma}=\left(\begin{array}{cc}
		\sigma_{U}^{2} &  \\
	 & \sigma_{V}^{2}
	\end{array}\right),
\end{equation}
by simplistically assume an independent relation between $U(t)$ and $V(t)$.
  $\sigma_{U},\sigma_{V}$ are set as a constant value that controls the variation of the moving process, and the mean $\mu_{U},\mu_{V}$ are set as equal to the motion state at the previous moment, i.e.,  $[\mu_{U}(t),\mu_{V}(t)]^T = [U(t-1), V(t-1)]^T $, by which a random motion process while keeping  smoothness is simulated for each artificial occluder.
 
The initial moving state $\mathbf{S}(0)$ is also defined as a random variable, where moving speed $|\mathbf{S}(0)| $ is sampled from a 1-D 
Gaussian distributions $ \mathcal {N}(	\mu,\sigma) $, and moving angle is sampled from $ \mathbf {U}(	0,2\pi)$.
To fit the distribution of the dataset, $\mu$ is set to be equal to the statistical expectations of the moving speed across the whole training dataset and $\sigma = \frac{\mu}{3}$.

In summary, given a sequence of images  $\{I_t\}$, the process of dynamically simulating occlusion consists of the following steps:
1). Randomly cropping the image sequence.
2). Randomly selecting a keyframe $I_k$ from $\{I_t\}$.
3). Segmenting $I_k$ into sub-regions by the super-pixel segmentation algorithm.
4). Randomly extracting $N$ sub-regions $\{\Psi_n\}_{n=1}^{N}$ and creating new textures.
5). Randomly assigning $\{\Psi_n\}$ at any location in the first frame $I_0$ and initialing the moving state.
6). Starting from the $I_0$, the dynamic masking process is simulated based on Markov's principle. 

\begin{figure}[!t]
	\centering
	\includegraphics[width=3.5in]{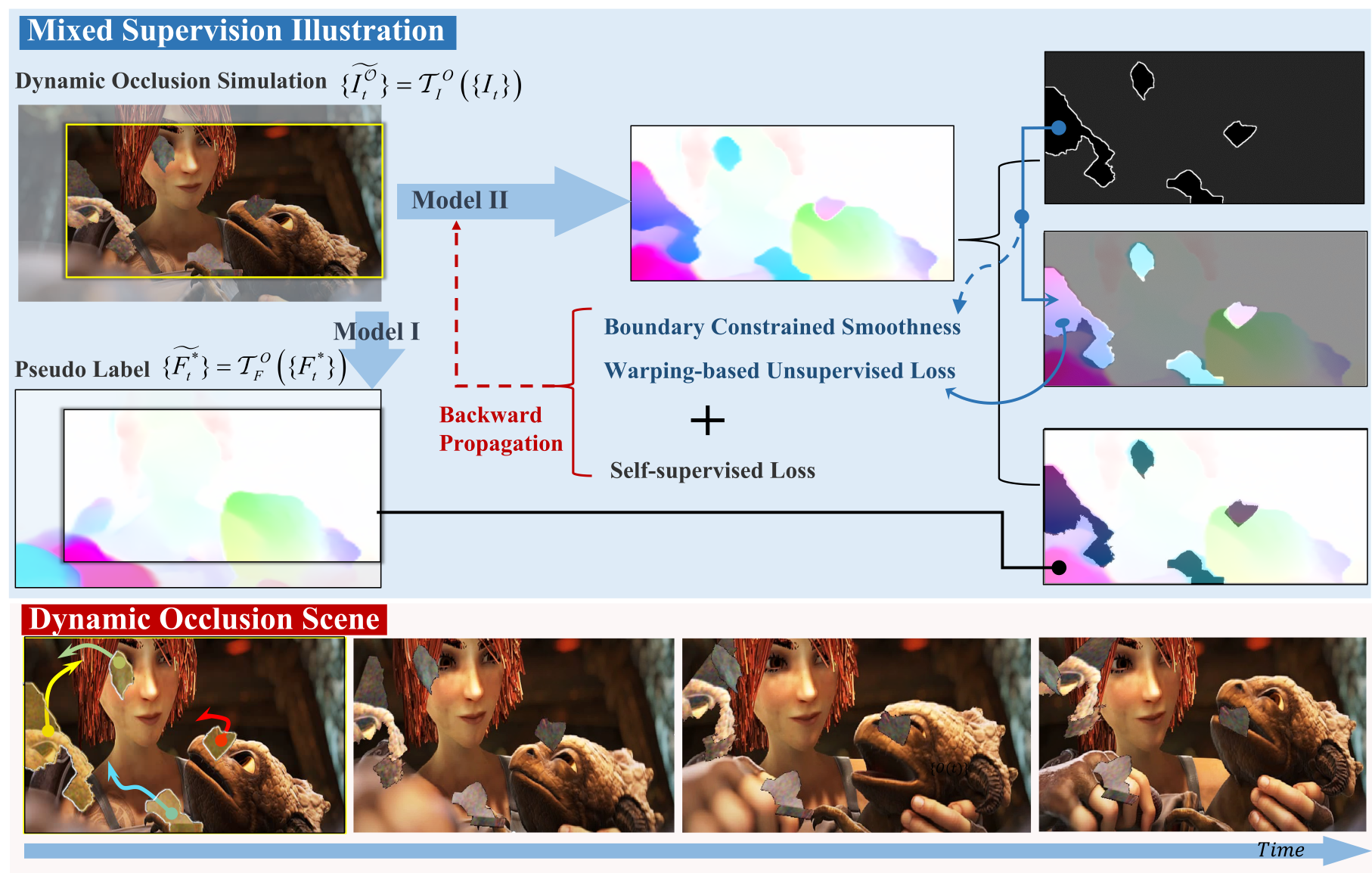}
		\vspace{-0.3cm} %
	\caption{\textit{\textbf{The illustration of mixed Supervision strategy and dynamic occlusion scene.}} To simulate natural object motion scenes, we utilize the superpixels algorithm to extract sub-objects of the image to act as artificial ``occluders'' with natural texture, and Markov processes are subsequently used to simulate the smoothness of the object's random motion,  the effect of which is shown in the bottom of the figure, where the occluders are highlighted by the yellow areas. Using the known prior of occlusion masks, we combine self-supervised learning and unsupervised learning to design a mixed supervised strategy, as shown in the blue box above. The details are described in subsection \ref{s5a}.}
		\vspace{-0.4cm} %
	\label{f3}
\end{figure}

In addition to translations, we also simulate more complex motions, such as twists, rotations, and a series of affine transformations, which basically follow the same Markov process as above to maintain the smoothness and randomness of the moving process.
For simplicity, we denote the transformation of the above process as $\mathcal{T}_I^O(\cdot): \{I_t\} \longmapsto \{\widetilde{I^{\mathcal{O}}_t}\}$ for image sequence, and $\mathcal{T}_F^O(\cdot): \{F^*_t\}\longmapsto\{\widetilde{ F^*_t}\}$ for pseudo label, correspondingly.
The visualization results of $\mathcal{T}_I^O(\cdot)$ and $\mathcal{T}_F^O(\cdot)$ are shown in the Fig. \ref{f3}.

\begin{figure*}[!t]
	\centering
	\vspace{-0.5cm} %
	\includegraphics[width=7in]{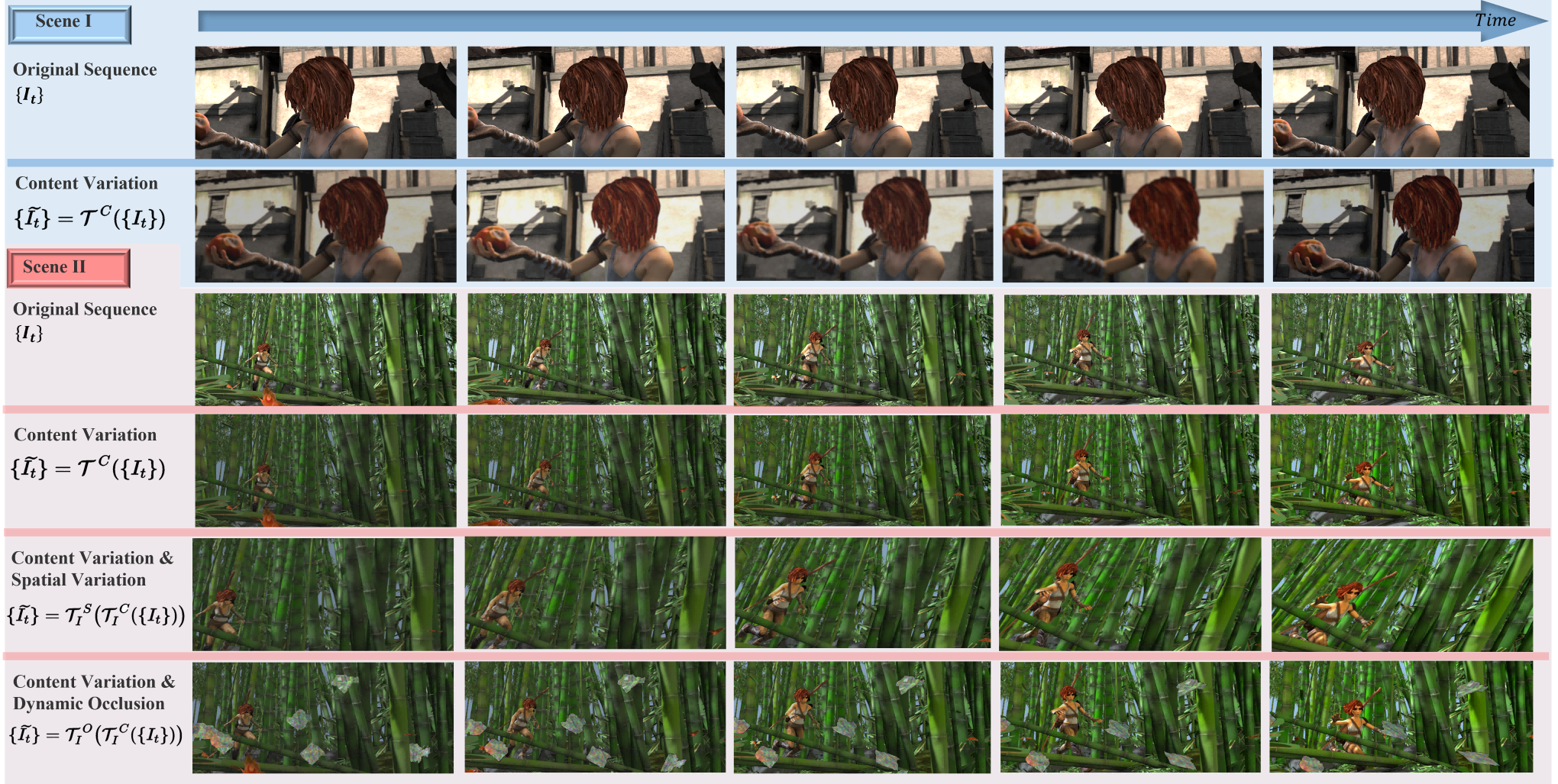}
		\vspace{-0.2cm} %
	\caption{\textbf{\textit{Qualitative presentation of simulated temporal dynamic variation scenes,}} including content variation scenes, spatial variation scenes, and dynamic occlusion scenes. Across the sequence, please note the dynamic changes in color and illumination, the dynamic shifts in spatial position, and the dynamic movement of occluders. Through self-supervised distillation, these dynamic scenes are well adapted by the model.  See Section \ref{s5} for more details about Dynamic Occlusion \ref{s5a}, and Content \& Spatial Variation \ref{s5b}.}
		\vspace{-0.4cm} %
	\label{f4}
\end{figure*}
\subsubsection{Mixed Supervision Strategy }
Two types of supervision strategies are proposed for the occlusion enhancer, which are Sparse Supervision and Mixed Supervision, respectively, both of which need to utilize pseudo label $\widetilde{F^*_t}$ for self-supervision.
As shown in Fig. \ref{f3}, at the moment $t$,  each occluder $\{\Psi_n^t\}_{n=1}^{N}$ could be considered as the closest object to the lens as the highest level of occlusion.  It is not difficult to generate the corresponding occlusion mask ${O}_t$ that follows:

\begin{equation}
	O_t(\mathbf{p})= \begin{cases}
		1, & \mathbf{p}\notin  \Omega_{\Psi}
		\\
		0, & \mathbf{p}\in   \Omega_{\Psi}
	\end{cases} 
\end{equation}
where $\Omega_{\Psi} = \{\Psi_1^t \cup \Psi_2^t \cup \dots  \cup\Psi _N^t  \}$.
  Based on the occlusion mask, the sparse supervision avoids the loss calculation of the occluder itself and only the supervise the regions without the occluder via pseudo label:

\begin{equation}
	\ell_{1}(t) \sim \mathbf{E}_{\mathbf{p}}\left(\mathbf{C}\left(\widetilde{F^*_t}(\mathbf{p}) , F_t(\mathbf{p})\right) \odot O_t(\mathbf{p})\right).
\end{equation}

The principle of sparse supervision here is only on reconstructing the occluded region instead of the occluder per se. In contrast, the mixed supervision strategy combines the self-supervised loss $\ell_{1}$  and the unsupervised loss $\ell_2$, which considers all regions of the flow field simultaneously. The self-supervised loss  is used for the $\mathbf{p} \notin \Omega_{\Psi}$, while the unsupervised loss based on image warping is available for the regions  $\mathbf{p} \in \Omega_{\Psi}$, which can be expressed by the following equation:

\begin{equation}
\ell_{2}(t) \sim {\operatorname{ SSIM}} \big( \mathcal{W}_t (I_{t+1}),I_t \big) \odot \left(1-O_t\right) +\ell_{sm},
\end{equation}
where $ {\rm SSIM}(\cdot) $ only operate on regions with occluder by multiplying with mask $1-O_t$.
Since the artificial occluders have rich texture information with time-invariant property, it is more suitable for using the pixel-wise similarity evaluation function, and the SSIM loss function is adopted in this case. 
Here $\ell_{sm}$ is similar to equation \ref{e1}, but the image gradient-based attenuation term is modified by replacing image $I_t$ with occlusion mask $O_t$, which will restrict the smoothing regularization within a sub-region of each occluder. 

Finally, the mixed supervision strategy of DOE is expressed as:
\begin{equation}
	\label{e15}
	\ell_{doe}(t) = \ell_{1}(t)+ \ell_{2}(t).
\end{equation}
It is worth noting that the occluder produces re-occlusion in the moving process, which also affects the warp-based unsupervised loss function. Therefore, we add a collision detection mechanism to the mixed supervision, which can avoid collisions between occluders during the random motion process, i.e., the whole moving process should satisfy $ \{\Psi_1^t \cap \Psi_2^t \cap \dots  \cap\Psi _N^t  \} = \emptyset$.

The sparse loss and mixed loss strategies are separately evaluated in the ablation study, in which the latter shows a better performance. Please check Table \ref{t8} for details.

\subsection{Spatial \& Content Variation Enhancer}
\label{s5b}

A series of spatial transformations are implemented in the SVE, including random cropping of sequences, rotation, horizontal and vertical flipping, and more complex transformations such as thin-plate-spline or CPAB transformations\cite{CPAB}, which can actually be regarded as spatial data augmentation.
Furthermore, we extend the spatial transformation to continuous time series and simulate various spatial variation scenarios across time.
In natural image sequences, there are scenes of spatial jitter and distortion due to environmental vibrations, camera shake, etc. Several such challenging scenes are rendered as examples in the Sintel Final dataset\cite{sintel}, which we generalize as the spatial variation. Accordingly,  a series of random affine transformations along the sequence are used to simulate the variation of the scene, including the dynamic rotation, distortion, scaling, and translation of objects. 
Specifically, these spatial transformations (except image cropping) on an image are just the shift of the spatial position of pixels, which act on the whole sequence  $\mathcal{T}^S_I: \{I_t\}\longmapsto \{\widetilde{I_t^{\mathcal{S}}}\}$, also can be represented as:
\begin{equation}
	\{\widetilde{I}_{t}^{\mathcal{S}}(\mathbf{p})\}=\big\{I_{t}\big(\tau_{\theta_t}(\mathbf{p})\big)\big\},
\end{equation}
where $\tau_{\theta_t}(\cdot)$ is the transformation of pixel coordinates at image $I_t$ with transform parameters of $\theta_t$.
To keep the consistency between transformed scene and pseudo label $F^*_t$, the optical flow needs to undergo different transformations, as different transformation parameters applied on $I_t $ and $ I_{t+1}$ will lead to a variation in optical flow field. In this case, one should first track the flow variation between $\{\tau_{\theta_t}, \tau_{\theta_{t+1}}\}$ via inverse-affine transformation, which subsequently be superimposed by the offset of original flow field $F^*_t$, and finally $\tau_{\theta_t}(\cdot)$ is used to keep spatially consistent with $\widetilde{I}_{t}$. The whole process $\mathcal{T}^S_F: \{F_t^*\}\longmapsto \{\widetilde{F^*_t}\}$ could be represented as:
\begin{equation}
	\left\{\begin{array}{l}
		F_{\text{new}}(\mathbf{p})=\tau_{\theta_{t+1}}^{-1}\left(\mathbf{p}+F^*_{t}(\mathbf{p})\right)-\tau^{-1}_{\theta_t}(\mathbf{p}) \\
	\widetilde{F_t^*}(\mathbf{p})=F_{\text{new}}\left(\tau_{\theta_t}(\mathbf{p})\right)
	\end{array}\right.
\end{equation}
Given pseudo label $\widetilde{F_t^*}(\mathbf{p})$, the loss function based on self-supervised learning of spatial variation enhancer can be expressed as:

\begin{equation}
	\label{e18}
	\ell_{sve}(t) \sim \mathbf{E}_{\mathbf{p}}\left(\mathbf{C}\left(\widetilde{F^*_t}(\mathbf{p}) , F_t^{\mathcal{S}}(\mathbf{p})\right) \right)
\end{equation}

Similar to the SVE, we introduce various transformations to the image content in the CVE. First, we introduce overall color jitter following the basic data augmentation principles\cite{arflow}, including random brightness, random saturation, random hue,  gamma transformation, etc. In addition, Gaussian blur and noise are randomly practiced on the whole image sequence to make the scene more challenging.

In addition to static augmentation, there are considerable content variation scenes in natural environments, such as changes in brightness and hue of objects due to the variation of illumination setting across time. Besides, various blurring phenomena that occur during motion, such as motion blur, defocus blur, etc., can also yield pixel mismatch across frames, posing a massive challenge for unsupervised optical flow estimation.
We simulate such scenarios in sequences with two types of variation, one with monotonically increasing or decreasing illumination, saturation, and hue along time, and the other with random jitter across times, which propel the network to focus on structural information of the object instead of the pixel differences. Similarly, we simulate a variety of dynamic blur scenarios, including box blur and Gaussian blur with random kernel size, defocus blur with random kernel size and location, linear motion blur with random orientation and kernel size, as well as the more complicated Point Spread Function (PSF) Blur \cite{psf}. 
Each of these blurs is assigned with random parameters in processing the sequence, accompanied by dynamic Gaussian noise that changes over time.

Specifically, the content transformation of a sequence of images is denoted as $\mathcal{T}_I^C(\cdot): \{I_t\}\longmapsto \{\widetilde{I}^{\mathcal{C}}_t\}$. As a relatively simple case,  $\mathcal{T}_I^C(\cdot)$ does not change the location of pixels nor introduce new occlusion. Therefore, there is no extra transform for the corresponding pseudo label, and the self-supervised loss of content variation training enhancer can be expressed as:

\begin{equation}
	\label{e19}
	\ell_{cve}(t) \sim \mathbf{E}_{\mathbf{p}}\Big(\mathbf{C}\big({F^*_t}(\mathbf{p}) , F_t^{\mathcal{C}}(\mathbf{p})\big) \Big)
\end{equation}

The qualitative illustration of three transformations  $\mathcal{T}_I^O(\cdot)$, $\mathcal{T}_I^S(\cdot)$, and $\mathcal{T}_I^C(\cdot)$ is shown in Fig. \ref{f4}.

\begin{figure*}[!t]
	\centering
		\vspace{-0.5cm} %
	\includegraphics[width=6.5in]{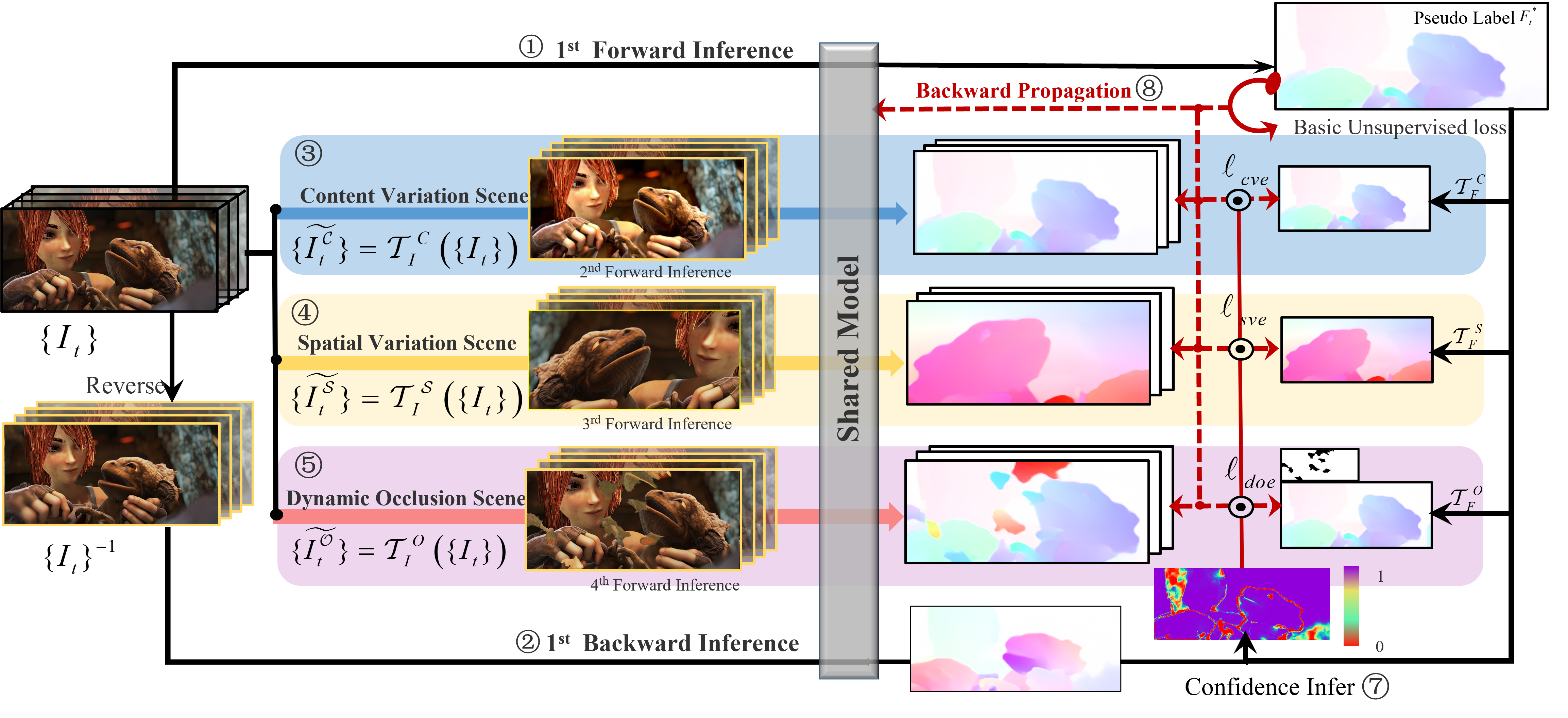}
	\vspace{-0.3cm} %
	\caption{\textit{\textbf{The complete training pipeline based on self-supervised learning. }}Three of the dynamic training enhancers expand the original sequence into content variant scenes, spatial variant scenes, and dynamic occlusion scenes, respectively. Four forward and one backward inference are performed in each iteration, where the first forward and backward inferences are used for the basic unsupervised training of the original scenes while computing the confidence mask. The subsequent triple forward inferences are carried out in extended scenes and back-propagated by a confidence-based self-supervised loss function. The entire process is operated on long dynamic sequences, allowing the network to understand the patterns of scene variation and occlusion in a dynamic environment. The execution order is marked from {\textcircled1} to {\textcircled9}, where  {\textcircled1} and {\textcircled2} corresponds to subsection \ref{sc12}; {\textcircled3},{\textcircled4} and {\textcircled5} to \ref{s5a}, and \ref{s5b} respectively; the final confidence inference {\textcircled7} and self-supervised distillation {\textcircled8} correspond to subsection \ref{s5c}.}
	\vspace{-0.4cm} %
	\label{f5}
\end{figure*}

\subsection{Self-supervised Distillation}
\label{s5c}
In self-supervised learning, a teacher model is required to generate pseudo-labels to supervise the predictions of the student model. However, the process can be simplified to online learning based on a single model that acts as both the teacher and the student model.

We adopt the strategy of ARFlow\cite{arflow} by running an extra forward pass with each transformed scene, along with using transformed pseudo label as the self-supervision signal. Specifically, we first expand the scenes using the previously defined transformations  $\{\mathcal{T}_I^O(\cdot), \mathcal{T}_I^S(\cdot),\mathcal{T}_I^C(\cdot)\}$ to obtain the three scenes $\{\widetilde{I^{\mathcal{O}}_t}\}, \{\widetilde{I^{\mathcal{S}}_t}\}, \{\widetilde{I^{\mathcal{C}}_t}\}\}$ based on the original scene sequence $\{I_t\}$.
The training process will perform four times forward inferences, in which the first one will generate the optical flow pseudo-label $\{F_t^*\}$ of the original scene, and the following three produce the optical flow of the dynamic occlusion scene, spatial variation scene, and content variation scene, respectively, which need to undergo corresponding flow transformation $\{\mathcal{T}_F^O(\cdot), \mathcal{T}_F^S(\cdot),\mathcal{T}_F^C(\cdot)\}$ to keep the consistency with the image transformation. By combining the loss functions in Eq. \ref{e7}, Eq. \ref{e15}, Eq. \ref{e18}, and Eq. \ref{e19}, the final loss function for self-supervised distillation can be formalized as:

\begin{equation}
	\label{e20}
		\begin{aligned}
		\mathcal{L}_{self}\sim\frac{1}{N-1} \sum_{t=1}^{N-1} \underbrace{\mathcal{L}_t}_{\text{\tiny 1st Infer}} + \Lambda\cdot \big[\underbrace{\ell_{doe}(t), \ell_{sve}(t) ,\ell_{cve}(t)}_{\text{\tiny2nd, 3rd, 4th Infer}}\big].
	\end{aligned}	
\end{equation}
Where $\Lambda = [\lambda_{3},\lambda_{4},\lambda_{5}]^T$ is the combined weight parameters for self-supervised loss.
By back-propagating $\mathcal{L}_{self}$, one can achieve both the basic unsupervised optical flow learning and self-supervised distillation in a single iteration.

In the process of self-supervision, we add a confidence propagation mechanism with reference to DDFlow\cite{ddflow}.
Based on the forward-backward consistency check, we define a confidence measure that describes the network's confidence for all spatial locations.
For a moment $t$, the forward flow $F_t $ and backward flow $ F_{t+1}^{-1}$ should satisfy:  $ F_t =- F_{t+1}^{-1}(\mathbf{p}+F_t(\mathbf{p})) $, or $| F_{t} + F_{t+1}^{-1}(\mathbf{p}+F_t(\mathbf{p}))| = 0 $. This equation does not hold in regions where either there is an error in the prediction or where the occlusion happened, both of which are defined as low confidence region in this case.
Based on this definition, the confidence mask could be generated as:

\begin{equation}
O_{conf} = {\rm MAP}\Bigg(\frac{\Big| F_t + F_t^{-1}(\mathbf{p}+F_t\big(\mathbf{p})\big)\Big|^2}{\big(|F_t|^2 + |F^{-1}_t|^2\big) \times \delta}\Bigg).
\end{equation}
Where the denominator is the weight decay based on the optical flow displacement; $\delta = 0.01$ is a scale factor, and $\rm MAP(\cdot) $is a nonlinear exponential mapping function that compresses the confidence level between $(0, 1)$. Besides, we also assign zero confidence to locations with excessive displacement based on prior knowledge.

The overall pipeline of the proposed self-supervised distillation training process is illustrated in Fig \ref{f5}.

\section{Experiment}
\label{s6}
We carry out comprehensive experiments on several  well-established datasets, including MPI-Sintel\cite{sintel}, KITTI 2012\cite{kitti2012}, KITTI 2015\cite{kitti2015}, FlyingChair\cite{flownet}, ChairSDHom\cite{flownet2}. In the experiment,  MPI-Sintel and KITTI are used to validate the performance of the proposed model against a range of state-of-the-art methods. The FlyingChair dataset is used for model pre-training, and the ChairSDHom is used for validating cross-datasets generalizability. 
For simplicity, our approach is denoted as \textbf{ULDENet} (\textbf{U}nsupervised \textbf{L}earning in \textbf{D}ynamic \textbf{E}nvironments) in comparison.

\begin{table}[!t]
	
	\renewcommand{\arraystretch}{1.1}
	\caption{\scriptsize \textbf{Comparison with previous methods on Sintel Benchmark}. We use the average EPE error (the lower the better) as evaluation metric for  the Sintel datasets. Missing entries `-' indicates that the result is not reported in the compared paper, and (·) indicates that the testing images are used during unsupervised training. The best unsupervised and supervised results are bolded, respectively. }
	\label{t1}
	\centering
	\setlength{\tabcolsep}{2.1mm}{
			\begin{tabular}{m{0cm}<{\centering}c|ccccc}
				\hline
				\cline{1-7}
				\multicolumn{2}{c}{\multirow{2}{*}{Method}} & \multicolumn{2}{c}{Sintel Train} & \multicolumn{2}{c}{Sintel Test} & \multirow{2}{*}{Param.} \\ 
				\cline{3-6}
				\multicolumn{2}{l}{}                        & Clean           & Final          & Clean          & Final          &                       \\ \hline
				\multirow{7}{*}{\rotatebox{90} { supervised}}       & FlowNetS\cite{flownet}          &   4.50              &         5.45       &    7.42            &     8.43           &               32.07M          \\ 
				& LiteFlowNet\cite{liteflownet}       &          (1.64)      &     (2.23)           &   4.86            &           6.09     &              5.37M           \\
				& PWCNet \cite{pwcnet}           &       (2.02)          &       (2.08)         &      4.39          &   5.04             &        8.75M                 \\
				& IRR-PWC \cite{hur2019iterative}          &        (1.92)         &       (2.51)         &      3.84          &        4.58        &        6.36M                 \\
				& SelFlow-ft \cite{selflow}          &        (1.68)         &   (1.77)             &      3.74          &        4.26        &       4.79M                  \\
				& STARFlow  \cite{starflow}        &           (2.10)      &     (3.49)           &       2.72         &         3.71       &         \bfseries   4.77M              \\
				& RAFT   \cite{raft}           &          \bfseries  (0.77)     &           \bfseries(1.20)    &        \bfseries   2.08     &      \bfseries   3.41        &          5.26M               \\ \hline
				\multirow{12}{*}{\rotatebox{90} {unsupervised}}     
				& UnFlow \cite{censusloss}           &      -           &       (7.91)         &       9.38         &     10.22           &         116.58M                \\ 
				& OAFlow  \cite{wang2018occlusion}    &      (4.03)         &      (5.95)          &         7.95       &        9.15        &   5.12M                      \\
				& MFOccFlow \cite{janai2018unsupervised}        &       (3.89)         &       (5.52 )        &       7.23         &         8.81       &   12.21M                      \\
				& DDFlow \cite{ddflow}           &       (2.92)        &      (3.98)          &       6.18         &       7.40         &      4.27M                   \\
				& SelFlow \cite{selflow}          &        (2.88)        &      (3.87)          &        6.56        &         6.57       &      4.79  M                 \\
				& ARFlow \cite{arflow}           &        (2.79)        &       (3.73)         &       4.78         &        5.89        &      \bfseries  2.24 M                 \\
				& ARFlow-MV\cite{arflow}         &       (2.73)    &        (3.69)        &         4.49       &      5.67          &        2.37M                 \\
				& SimFlow\cite{2020deepfeaturematching}             &        (2.86)      &        (3.57)        &         5.92       &          6.92      &     9.70M      \\
				&  UFlow\cite{uflow}          &       (2.50)       &       (3.39)         &   5.21             &       6.50         &             -            \\
				& UPFlow\cite{luo2021upflow}            &    (2.33)         &       (2.67)         &      5.24          &         6.50       &           3.49M              \\
				& OIFlow\cite{oiflow}            &      (2.44)         &      (3.35)        &     4.26         &        5.71       &          5.26M              \\
				
				& \bfseries Ours (ULDENet)         &    \bfseries  (2.20)          &  \bfseries(2.65)              &    \bfseries  3.50           &  \bfseries    4.70          &                2.50M        \\
				\hline
				\cline{1-7}
				
			\end{tabular}
		\vspace{-0.5cm} %
		}
	\end{table}
	
	\begin{table}[!t]
		
		\renewcommand{\arraystretch}{1.1}
		\caption{\scriptsize \textbf{Comparison with previous methods on KITTI Benchmark}. We use the average EPE error (the lower the better) as evaluation metric for the KITTI 2012 datasets, F1 measurement for KITTI 2015 test benchmark. Missing entries `-' indicates that the result is not reported in the compared paper, and (·) indicates that the testing images are used during unsupervised training. The best unsupervised and supervised results are bolded, respectively. }
		\label{t2}
		\centering
		\setlength{\tabcolsep}{1.7mm}{
				\begin{tabular}{m{0cm}<{\centering}c|ccccc}
					\hline
					\cline{1-7}
					\multicolumn{2}{c}{\multirow{2}{*}{Method}} & \multicolumn{2}{c}{KITTI 2012} & \multicolumn{2}{c}{KITTI 2015} & \multirow{2}{*}{Param.} \\ 
					\cline{3-6}
					\multicolumn{2}{c}{}                        & Training           & Test          &  Training          & Test          &                       \\ \hline
					\multirow{7}{*}{\rotatebox{90} { supervised}}      
					& FlowNet2-ft\cite{flownet2}      &          1.28       &    1.8            &         2.30       &         11.48\%       &       162.5M                  \\
					& PWCNet\cite{pwcnet}         &           1.45      &       1.7         &      2.16          &         9.60\%       &          8.75M               \\
					& IRR-PWC \cite{hur2019iterative}         &           -      &       -         &       1.63         &         7.65\%       &            6.36M             \\
					& SelFlow-ft\cite{selflow}         &     \bfseries   0.76          &    \bfseries 1.5            &        1.18        &        8.42\%        &           4.79M              \\
					& STARFlow\cite{starflow}        &          -       &          -      &     -           &         7.65\%       &         \bfseries  4.77M               \\
					& RAFT\cite{raft}       &         -        &        -        &   \bfseries  0.64             &       \bfseries  5.27\%         &       5.26M                  \\ \hline
					\multirow{13}{*}{\rotatebox{90} {unsupervised}}      
					& UnFlow \cite{censusloss}          &       3.29          &        -        &    8.10            &     23.3\%           &              116.58M           \\ 
					& OAFlow \cite{janai2018unsupervised}     &        3.55         &       4.2         &      8.88          &        31.2\%        &                5.12M         \\
					& DDFlow\cite{ddflow}          &      2.35         &        3.0        &     5.72           &       14.29\%         &        4.27M                 \\
					& SelFlow\cite{selflow}          &           1.69      &       2.2         &      4.84          &         14.19\%       &       4.79M                  \\
					& EpiFlow \cite{epipolarflow} &(2.51) &3.4&(5.55) &16.95\% & 8.75M \\
					
					& NLFlow\cite{nlflow}           &    3.02             &    4.5            &   6.05            &     22.75\%           &           -              \\
					& ARFlow\cite{arflow}            &      1.44           &        1.8        &    2.85            &       11.8\%         &      \bfseries2.24M                   \\
					& ARFlow-MV\cite{arflow}       &       1.26          &         1.5       &         3.46       &         11.79\%       &         2.37M               \\
					& UFlow\cite{uflow}            &       1.68          &       1.9         &        2.71        &           11.13\%     &         -                \\
					& UPFlow \cite{luo2021upflow}          &       1.27          &    1.4            &     2.45           &         9.38\%       &        3.49M                 \\
					
						& OIFlow \cite{oiflow}          &       1.33          &    1.6            &     2.57           &         9.81\%       &        5.26M                 \\
					& \bfseries Ours (ULDENet)     &     \bfseries   (1.15)         &         \bfseries1.3       &           \bfseries (2.23)      &    \bfseries  9.13\%       &                  2.50M      \\
					\hline
					\cline{1-7}
				\end{tabular}
			}
		\vspace{-0.5cm} %
		\end{table}

\subsection{Implementation Details}
\label{s6b}
According to Eq. \ref{e7} and Eq. \ref{e20}, we  set the loss function weights $\{\lambda_1, \lambda_2, \lambda_3, \lambda_4 ,\lambda_5\}$ to $\{50,0.005,0.3,0.3,0.3\}$ for Sintel dataset and $\{75,0.001,0.2,0.2,0.2\}$ for KITTI dataset, respectively.
We implement our method using Pytorch on a workstation with four paralleled RTX A6000 GPUs under CUDA 10.1.
All models are trained by Adam optimizer \cite{adam} with $\beta_1 = 0.9$, $ \beta_2 = 0.99$. We first pre-train the model on FlyingChair with batch size of 8 and a learning rate of $\rm 1e^{-4}$, followed by two stages of formal training on the Sintel or KITTI dataset. In the first stage, we use the basic unsupervised loss function with a sequence length of 6, batch size of 4, and learning rate of $1e^{-4}$ for 100 epochs of training. In the second stage, we perform 50 epochs of intensive training using self-supervised distillation, where the sequence length is set to 8 per iteration, the batch size to 4, and the learning rate is set to $\rm 5e^{-5}$ for Sintel and $\rm 8e^{-5}$ for KITTI, respectively. The gradient clipping technology is practiced in all training stage for faster convergence.

For image preprocessing, the images from the Sintel dataset are cropped to $384\times832$ and inferred at the original size in the validation; the images in the KITTI dataset are resized to  $256\times832$ for both training and validation. We also adopt random image horizontal flipping and sequence reversion for dataset augmentation.

In training, we mix Clean, Final, and Albedo scenes from the Sintel training set and extracted them randomly for training. As for training on the KITTI dataset, we employ its multi-view extension version for sequence training. It is worth to note we only touch the images of the training set in all training procedures, and the flow ground-truth is only used for validation. The standard average endpoint error (EPE) and the percentage of erroneous pixels (F1) are used as the evaluation metric of optical flow. In the validation of the Sintel dataset, we dynamically input all the frames of each scene into the model at once and validate the results of each frame; In the validation of the KITTI dataset, we input frames from 1th to 11th into the model and only validate the results of frame 10th.

\begin{figure*}[!t]
	\vspace{-0.5cm} %
	\centering
	\includegraphics[width=7.0in]{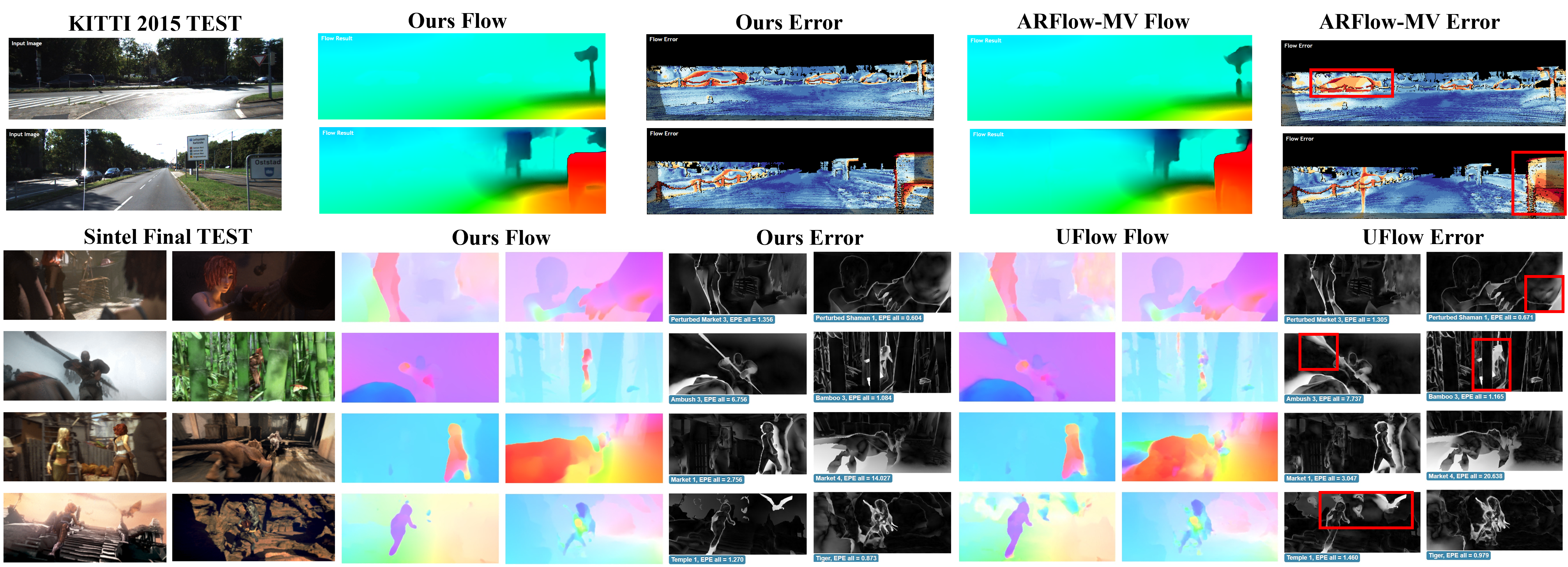}
	\caption{\textit{\textbf{Visualization of results on Sintel and KITTI datasets.}} We qualitatively compare the results with ARFlow-MV\cite{arflow} and UFlow\cite{uflow}, where the red boxed area marks our dominant region. Please zoom in to see the detailed information, and the quantitative comparisons are shown in Table \ref{t1} and Table \ref{t2}. All results in this figure were generated by the official server on the test set without subjective picking.}
	\label{fs}
	\vspace{-0.4cm} %
\end{figure*}

\subsection{Comparison on MPI-Sintel and KITTI Benchmarks}
We compare the proposed method with both supervised and unsupervised approaches on the Sintel Clean and Final dataset. The test set result is first generated and then verified by submitting it to the Sintel official server. In Table \ref{t1}, our method achieves the best accuracy among all compared unsupervised methods on both training and test datasets with fairly low parameter overhead. In particular, we achieved a significant lead in the Sintel test set, where the result of the Clean scene outperforms the previous best approach OIFlow\cite{oiflow} by an 18\% decrease on the EPE metric with only 48\% memory overhead. Moreover, we improve the previous best method ARflow-MV from EPE=5.67 to EPE=4.70 with a 17\% error reduction on the Sintel Final test set. 

Compared to supervised methods, ULDENet is comparable to SelFlow-ft\cite{selflow} on the Sintel test benchmark and to the 2020 work STARFlow\cite{starflow} on the training set, with the advantages of lower parameters overhead.


We evaluate the proposed method both on KITTI 2012 and KITTI 2015 benchmarks. The quantitative result could be checked in Table \ref{t2}.
Compared with other unsupervised learning methods, ULDENet achieves the best results on both KITTI 2012 and KITTI 2015 benchmarks, demonstrating the proposed strategy's efficiency. On the KITTI 2015 dataset, our method improves the previous best method UPFLow\cite{luo2021upflow} with EPE=2.45 to EPE=2.23, leading by 9\% error reduction with the advantage that the number of parameters is only 71\% of UPFlow. Moreover, on KITTI 2015 online evaluation, our method reduces the F1-all value of 11.1\% in UFlow\cite{uflow} to 9.1\%, with an 18.0\% error reduction. As for the KITTI 2012 evaluation, the results of ULDENet on the training set decrease the EPE=1.27 of UPFlow to EPE=1.15, with a reduction of 9 percent.

Compared with the supervised learning methods in the table, the performance of ULDENet is between that of PWCNet\cite{pwcnet} and SelFlow-ft\cite{selflow}, where the EPEs on both KITTI 2012 and 2015 datasets are ahead of PWCNet, but the number of the parameters is only 29\% of it.

The qualitative comparisons of the Sintel dataset and KITTI dataset are both visualized in Fig. \ref{fs}

\subsection{Ablation Study}
An extensive and rigorous ablation study is conducted to demonstrate the effectiveness of each technical component.
Sintel training set was subdivided into separate training and validation sets following the setting in \cite{arflow}. AEPE errors for all pixels (all), non-occluded pixels (NOC) and occluded pixels (OCC) are reported for quantitative comparison.

\subsubsection{Main Ablation}

A series of components is proposed in this work, including the multi-frame recurrent inference structure (MFI), temporal smoothness regularization (TSR), and dynamic training enhancer (DTE), which have been comprehensively verified on Sintel dataset. As summarized in Table \ref{t3}, the `Base' represents the baseline, i.e., the original lightweight PWCNet in a double-frames inference manner;
MF represents the model adopting a multi-frame training and inference way by adding the spatial-temporal dual recurrent block, which has a comprehensive improvement for performance and increases the number of parameters by 0.4M. After introducing TSR, the model becomes more robust for the occluded region. Attributing that TSR can provide a more reliable reference for the occluded region via the motion prior to the adjacent frames, the EPE of OCC regions on the clean dataset reduces from 19.36 to 18.74. 
Furthermore, DTE significantly improves the network's performance without additional memory and computational overhead. Quantitatively, the network performance improves from EPE=2.17 to EPE=1.75 by DTE, with a 21\% decrease in EPE for the OCC region.
Integrating all the strategies, we eventually improved the performance from EPE = 2.23 to EPE = 1.67 for the Clean dataset, and from EPE = 3.23 to EPE = 2.76 on the Final dataset, reducing the end-point error by 20\% on average.


\begin{table}[!t]
	\vspace{-0.5cm} %
	\renewcommand{\arraystretch}{1.1}
	\caption{\scriptsize \textbf{Main ablation study on different combinations of the proposed components.} \textbf{Base:} the baseline of original lightweight PWCNet in a double-frames inference manner. \textbf{MF:} multi-frame training and inference structure. \textbf{TSR:} temporal smoothness regularization. \textbf{DTE:} dynamic training enhancer. }
	\label{t3}
	\centering
	\setlength{\tabcolsep}{1.0mm}{
	\begin{tabular}{cccc|cccccc|c}
		\hline \cline{1-11} 
		\multirow{2}{*}{Base} & \multirow{2}{*}{MF} & \multirow{2}{*}{TSR} & \multirow{2}{*}{DTE} & \multicolumn{3}{c}{Sintel Clean} & \multicolumn{3}{c}{Sintel Final} \vline& \multirow{2}{*}{Param.} \\ \cline{5-10}
		&                   &     &                 &ALL       & NOC             & OCC      &ALL      & NOC             & OCC            &                         \\ \hline 
	\checkmark 	&        & &               &                 2.23      &         0.97          &    19.71         &    3.23             &     1.80           & 23.69     &          2.10M              \\             
	\checkmark 	&           \checkmark        &     &  &        2.17      &         0.91          &    19.36     &    3.12   &1.74 &22.86   &   2.50M                     \\
	
	\checkmark 	&          \checkmark         &     \checkmark   &           &   2.10    &   0.88       & 18.74&    3.06   &       1.70          &    22.51           &     2.50M                    \\
	\checkmark 	&         \checkmark       &                     &   \checkmark                      &   1.75         &0.81 &   15.21   &   2.81     &  1.63     &  19.83   &     2.50M      \\ 
	\checkmark 	&        \checkmark        &      \checkmark     &    \checkmark                 &    \bfseries1.67            &  \bfseries0.77 &   \bfseries14.54   &     \bfseries2.76             &      \bfseries 1.61            &      \bfseries 19.37         &  2.50M         \\ 
	\hline \cline{1-11} 
	
	\end{tabular}
\vspace{-0.5cm} %
}
\end{table}

\subsubsection{Network Structure}

 Table \ref{t4} provides a concise comparison of the models with different structures. To separately demonstrate the efficiency of the proposed model structure, we use the same loss strategy as ARFlow, and DTE and TSR are not adopted in the comparison. The only difference in training is that our model is trained and inferred in a multi-frame environment. The results indicate that our network is substantially ahead of PWCNet\cite{pwcnet} in terms of performance and network parameters. 
Compared to the multi-frame approach ARFlow-MV\cite{arflow}, which extracts five adjacent frames before and after the current time, we only access the frames as inputs before the current time, allowing the ULDENet structure to fit well in a real-time task. Also, our method can handle sequences of arbitrary length, which can not be realized by most static multi-frame methods such as ARFlow-MV, SelFlow\cite{selflow}, MFOccFlow\cite{2018unsupervisedmultiframe}, etc. In terms of quantitative results, we significantly reduce the EPE of the network at the cost of a slight increase in memory overhead.

\begin{figure}[!t]
	\centering
	\vspace{-0.5cm} %
	\includegraphics[width=3.5in]{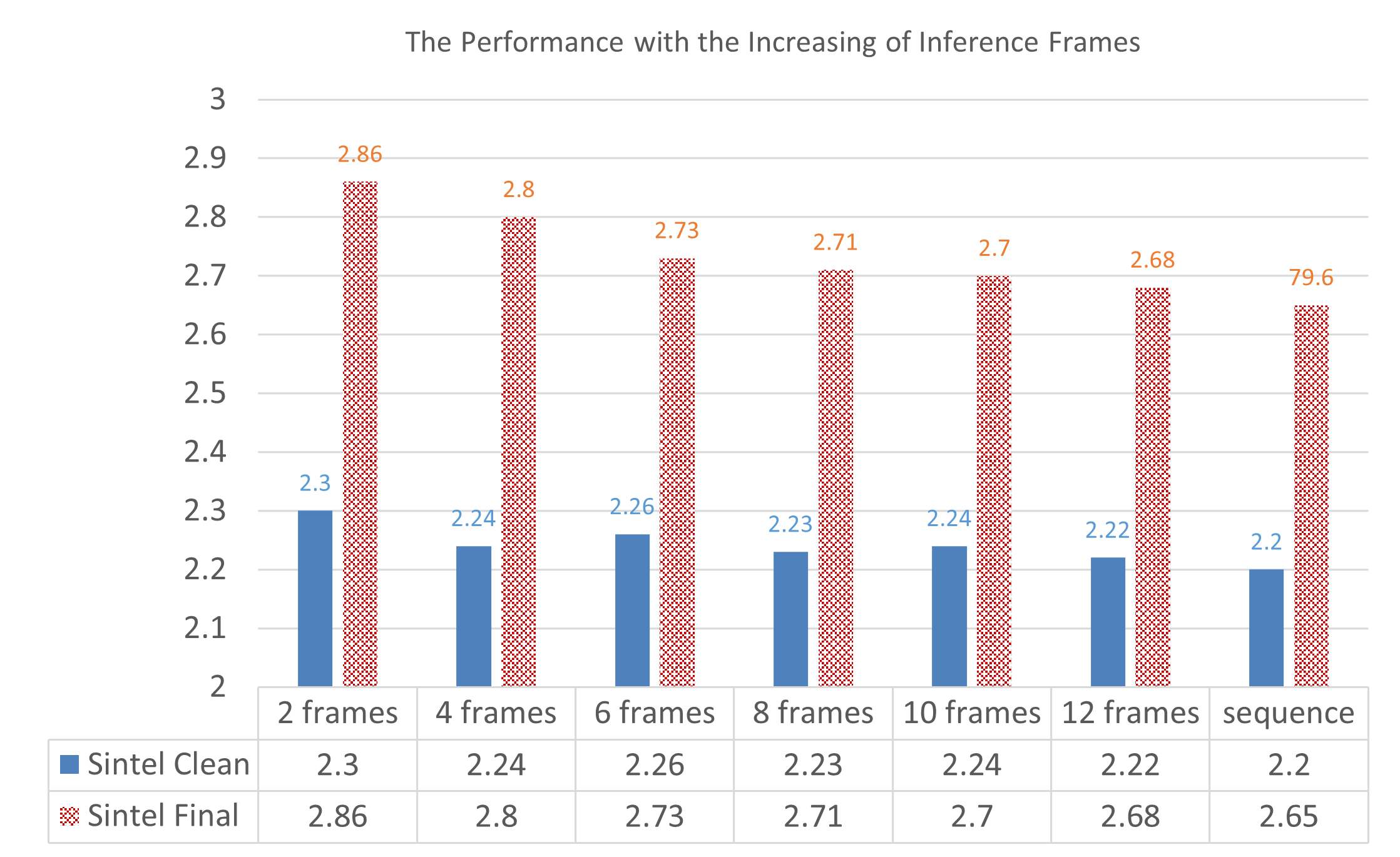}
	\vspace{-0.5cm} %
	\caption{\textbf{\textit{The network EPE performance w.r.t the increasing of inference frames.}} After multi-frame dynamic training, the proposed  model converges in sequences of arbitrary length. The \textbf{sequence }in the table indicate that the network utilizes all frames from the scene per inference. The model performance gradually improves as the inference frames rise. }
	\label{f6}
\end{figure}

It is worth mentioning that even though our model is trained in a dynamic environment with multiple frames, it still demonstrates good generalization in a two-frame inference approach after reaching convergence.
Specifically, we split each scene of the Sintel dataset into different frame patches to verify the impact of accuracy w.r.t inference frames of the network, and the results are shown in Fig. \ref{f6}.

Obviously, even if the inference is performed with only two frames, the EPE of our method for the Clean (EPE = 2.30) and Final  (EPE = 2.86) dataset is still substantially lower than that of the multi-frame method ARFlow-MV. Besides, the EPE of the model gradually decreases as the number of inference frames rises, where multi-frame inference brings more significant gain on the Final dataset with more complicated scenes,  suggesting ULDENet is more suited for working in  dynamic real-time environments with complex circumstances.

\begin{table}[!t]

	\renewcommand{\arraystretch}{1.1}
	\caption{\scriptsize \textbf{{Ablation study of our framework with multiple model architectures.}} AEPE in specific regions of the scene and the number of CNN parameters are reported.}
	\label{t4}
	\centering
	\setlength{\tabcolsep}{1.2mm}{
	\begin{tabular}{c|cccccc|c}
		\hline \cline{1-8}
	
		\multirow{2}{*}{Net Structure} & \multicolumn{3}{c}{Sintel Clean} & \multicolumn{3}{c}{Sintel Final }\vline & \multirow{2}{*}{Param.} \\ \cline{2-7}
		& ALL       & NOC       & OCC      & ALL       & NOC       & OCC      &                         \\ \hline
	PWCNet\cite{pwcnet}	&      2.48     &   1.19       &   21.71       &    3.47       &      1.98     &   25.19       &           8.75M              \\
	PWCNet-small\cite{pwcnet}	&     2.76      &    1.28       &  23.92        &    3.62       &    2.16       &   28.15       &          4.05M              \\
	ARFlow\cite{arflow}	&        2.30   &    1.08       &   20.00       &     3.19      &   1.84        &   22.77       &                \bfseries  2.24M       \\
	ARFlow-MV\cite{arflow}	&     2.24      &   1.04        &  19.60        & 3.18          &    1.86       &   \bfseries 22.36       &               2.37M          \\ \hline
	Ours-seq	&       \bfseries2.17      &     \bfseries 0.91       &    \bfseries 19.36       &      \bfseries  3.12     &     \bfseries   1.74     &    22.83       &                2.50M         \\
		\hline \cline{1-8}               
	\end{tabular}
}
\vspace{-0.5cm} %
\end{table}

\subsubsection{Temporal Smoothness Regularization}

\begin{table}[!t]
	
	\renewcommand{\arraystretch}{1.1}
	\caption{\scriptsize \textbf{Ablation study on weight of temporal smoothness loss $W_{tsm}$. }  AEPE in specific regions of the Sintel and KITTI datasets are reported.}
	\label{t5}
	\centering
	\setlength{\tabcolsep}{2.5mm}{
		\begin{tabular}{c|cccccc}
			\hline \cline{1-7}
			
			\multirow{2}{*}{$W_{tsm}$} & \multicolumn{3}{c}{Sintel Clean} & \multicolumn{3}{c}{KITTI-15} \\ \cline{2-7} 
			& ALL       & NOC       & OCC      & ALL      & NOC     & OCC     \\ \hline
			0                  &  2.17         &     0.91      &   19.36       &  2.65        &  \bfseries 1.92      &  5.86     \\
			0.01               &   2.15        &    0.92       &        19.07   &    \bfseries 2.52    &    1.94     &   \bfseries  5.07    \\
			0.05               & \bfseries  2.10        &   \bfseries   0.88     &     \bfseries   18.74  &      2.87    &      2.07   &    6.39     \\
			0.1                &      2.26     &   1.04        &       19.21    &     2.95    &      2.12   &  6.60       \\ 
			0.5                &      3.24     &     2.05      &      21.86    &    3.46       &      2.35    &   8.35   \\
			\hline \cline{1-7}               
		\end{tabular}
	}
	\vspace{-0.2cm} %
\end{table}

Based on the grid search method, we verified the effect of temporal smoothing regularization (TSR) on optical flow estimation with different weights, as summarized in Table \ref{t5}.
We used the ULDENet structure with the basic unsupervised loss strategy as a baseline,  and DTE was not operated in the validation. Both KITTI 2015 and Sintel datasets were included in the validation. From the results, large-scale TSR leads to degradation of the network performance, and accordingly, its weights should be restricted to a small range. Compared to NOC regions, the TSR brings more significant gains for estimating OCC regions,  as NOC regions are robustly governed by the photometric loss function, while mixed supervision of temporal smoothing regularization and spatial smoothing regularization is instead required for OCC regions.

According to Table \ref{t5}, The TSR weight is set to 0.01 for the KITTI dataset and 0.05 for the Sintel dataset, respectively.

\begin{table*}[!t]
	\vspace{-0.5cm} %
	\renewcommand{\arraystretch}{1.1}
	\caption{\scriptsize \textbf{Ablation study on different combinations of three training enhancers.} \textbf{Baseline:} ULDENet structure with basic multi-frame unsupervised training. \textbf{DOE:} dynamic occlusion training enhancer; \textbf{SVE:} spatial variation training enhancer; \textbf{CVE:} content variation training enhancer.}
	\label{t7}
	\centering
	\setlength{\tabcolsep}{2.5mm}{
		\begin{tabular}{cccc|cccccccccccc}
			\hline \cline{1-16}
			\multirow{2}{*}{Baseline}	 & 	\multirow{2}{*}{DOE} & \multirow{2}{*}{SVE} & \multirow{2}{*}{CVE} & \multicolumn{3}{c}{Sintel Clean} & \multicolumn{3}{c}{Sintel Final}                  & \multicolumn{3}{c}{KITTI-15}   &     \multicolumn{3}{c}{KITTI-12}                                  \\ 
			\cline{5-16} 
			&	&                      &                      & ALL                  & NOC & OCC & ALL & NOC                  & OCC                  & ALL                  & NOC                  & OCC    & ALL                  & NOC                  & OCC                  \\ \hline
			\checkmark	&& &&       2.17    & 0.91 &  19.36 & 3.12 & 1.74 & 22.86 &  2.65 &   1.92  &   5.86  &  1.42         &  0.92     & 4.32     \\
			\checkmark&\checkmark&&&    1.86    & 0.85    &    16.25 &  2.90     &  1.69    &  20.42  &  2.43    &1.76  &5.38 &1.28&0.85&  3.81  \\
			\checkmark	&&  \checkmark&& 1.96 &  0.75  &   17.87  & 2.98    &1.63     &    22.17      &   2.55         & 1.82    & 5.77  &1.35&0.86&  4.18 \\
			\checkmark	&	\checkmark	&&  \checkmark &    1.79   &  0.78 & 15.82    &  2.88   & 1.68  &   20.22   &  2.37       &  1.72       &  5.24        &1.24&0.83& 3.67    \\ 
			\checkmark	&& &  \checkmark & 1.92 &  0.79   &   17.23  &          2.94     &     1.66    &     21.36   & 2.50 &   1.86    &  5.60       &1.31&0.86&   3.96  \\ 
			\checkmark	&\checkmark	&   \checkmark  &\checkmark&1.75& 0.81   &  15.21    & 2.81      &  1.63   & 19.83    &  2.28   &     1.62   &    5.17  &    1.19    &0.78 & 3.59   \\
			\hline \cline{1-16}
		\end{tabular}
	}
	\vspace{-0.5cm} %
\end{table*}

\subsubsection{Dynamic Training Enhancer}

Based on self-supervised learning, the dynamic training enhancer (DTE) improves the model's generalization to the three specific scenes and thus enhances the model performance by a large margin. We separately validate the performance of its three components: Dynamic Occlusion Training Enhancer (DOE), Spatial Variation Training Enhancer (SVE), and Content Variation Training Enhancer (CVE), as well as the performance of their different combinations strategies.

Table \ref{t7} summarizes the ablation studies of DTE on the Sintel Clean, Sintel Final, KITTI 2012, and KITTI 2015 datasets, where the Baseline represents the ULDENet with the basic unsupervised loss combination,  and the temporal smoothing regularization is not used here. Quantitatively, the introduction of DOE alone reduces the EPE from 2.17 to 1.86, bringing the most considerable improvement of 14.3\%, and the following SVE and CVE obtain 10\% and 11.5\% performance gain, respectively. Combining the three can significantly reduce the EPE from 2.17 to 1.75, reducing the error of nearly 20 percent.
As the result of the dynamic and realistic simulation of occlusion, the DOE is most significant for improving the prediction of the OCC region, as diminishing the EPE from 19.36 to 16.25 on the Sintel Clean dataset. The CVE also slightly improves around the OCC region, as the dynamic noise and regional blurring can be considered similar to information loss from the occlusion. In addition, both CVE and SVE can reduce the prediction error of the network for regular NOC regions due to the simulation of multiple dynamic variation scenarios and self-supervised distillation.

The dynamic variation transformation and static image transformations are further validated in Table \ref{t8}, where the CT and ST are defined as identically performing the Content \& Spatial image transformations across all frame sequences, such as image flipping, Gaussian blur, etc., which are similar to the data augmentation method\cite{arflow}; Stc Occ is defined as image cropping and random noise occlusion of image subregions with the original unsupervised loss\cite{selflow}.
All transform hyperparameters are set to be consistent with the dynamic scene.
From the results, it is straightforward that dynamic variant scenes and dynamic occlusion have considerable performance advantages over static image augmentation, in which the proposed dynamic occlusion simulation with a mixed occlusion loss strategy (Dyc Occ-mixed) holds a gain of nearly 10\% over Stc Occ.

\begin{table}[!t]
	
	\renewcommand{\arraystretch}{1.1}
	\caption{\scriptsize  \textbf{Ablation study between dynamic scene variation and static data augmentation.}
		\textbf{Baseline:} ULDENet structure with basic multi-frame unsupervised training; \textbf{Stc Occ:} static random noise masking; \textbf{Dyc Occ-sparse / mix:} dynamic occlusion simulation with spares / mixed loss supervision. \textbf{Stc CT / ST:} static content transformation / spatial transformation. \textbf{Dyc CV / SV:} dynamic content variation / spatial variation.}
	\label{t8}
	\centering
	\setlength{\tabcolsep}{1.9mm}{
		
		\begin{tabular}{c|cccccc}
			\hline
			\cline{1-7}
			\multirow{2}{*}{Ehancer Strategy} & \multicolumn{3}{c}{Sintel Clean} & \multicolumn{3}{c}{Sintel Final}                  \\ \cline{2-7}
			& ALL                  & NOC & OCC & ALL & NOC                  & OCC                  \\ \hline
			Baseline                              & 2.17     &  0.91   &    19.36 & 3.12 &1.74&22.83\\\hline
			+Stc Occ                          &      2.06    & 0.90 &18.21&3.03&1.78&    21.15     \\
			+Dyc Occ-sparse                   &   1.90       &  0.89   &16.47     & 2.95    &       1.72             &       20.74               \\
			+Dyc Occ-mix                      &     \bfseries1.86     &  0.85    &    \bfseries16.25  & \bfseries2.90    &        1.69           &   \bfseries  20.42             \\ \hline
			+ Stc CT                          &     2.04     &  0.81   &  18.44   & 3.03    &        1.71              &       22.01               \\ 
			+ Dyc CV                          &     1.92     & 0.79   &  17.23   &  2.94   &          1.66            &  21.36                    \\ \hline
			+ Stc ST                          &  2.07   & 0.77    &  19.00  &3.08 & 1.73&22.44 \\
			+ Dyc SV                          &  1.96   &  \bfseries 0.75   &   17.87  & 2.98 & \bfseries1.63& 22.17\\
			\hline
			\cline{1-7}
		\end{tabular}
		
	}
\end{table}

\subsection{Generalization Validation}
\begin{table}[!t]
	\vspace{-0.2cm} %
	\renewcommand{\arraystretch}{1.1}
	\caption{ \scriptsize \textbf{Generalization Validation.} All models were fine-tuned on the Sintel dataset and validated cross-datasets.  }
	\label{t9}
	\centering
	\setlength{\tabcolsep}{1.7mm}{ 
		\begin{tabular}{c|cccc}
			\bottomrule
			Method       & ARFlow\cite{arflow} & PWCNet\cite{pwcnet} & RAFT\cite{raft} & \bfseries Ours          \\
			\hline
			Sintel Clean & (2.79)   & (1.86)   & \bfseries(0.74) &(2.20)         \\
			Sintel Final & (3.73)   & (2.31)   & \bfseries(1.21) & (2.65)         \\
			ChairSDHom   & 0.67   & 2.67   & 0.48 & \textbf{0.47} \\
			FlyingChair  &3.50   & 3.69   & \bfseries1.20 & 3.33 \\
			\bottomrule
		\end{tabular}
	}
\vspace{-0.3cm} %
\end{table}
		
 We validate the cross-dataset generalization performance of the proposed model and compare it with both supervised and unsupervised models. In Table \ref{t9}, we choose Sintel as the training set and FlyingChairs\cite{flyingchairs} and ChairSDHom\cite{flownet2} as the validation sets. Both supervised learning methods, RAFT\cite{raft}, PWCNet\cite{pwcnet}, and unsupervised learning method ARFlow\cite{arflow} are selected as comparisons.

It should be noted that both FlyingChair and ChairSDHom datasets only allow two frames per inference, which limits the capability of our model. However, the generalization performance of the ULDENet on the FlyingChair dataset is still higher than that of ARFlow and PWCNet.
In particular, our generalization performance on ChairSDHom is even higher than that of one of the most advanced supervised learning methods,  RAFT.

\begin{figure}[!t]
	\centering
	\includegraphics[width=3.0in]{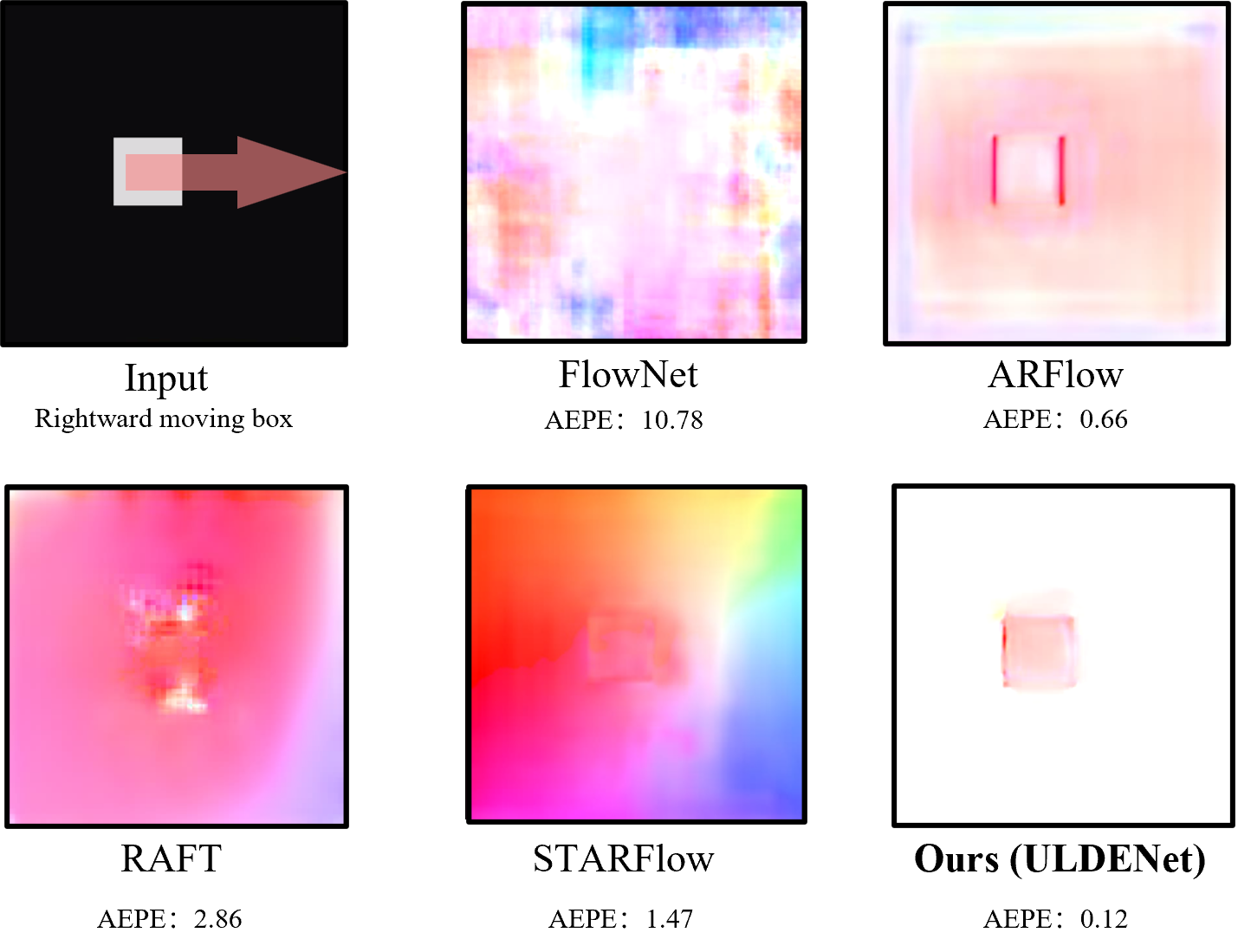}
	\vspace{-0.3cm} %
	\caption{\textit{\textbf{Demonstration of the generalization performance of the model in the non-textured stimuli.}} We synthesized 100 consecutive image frames with a constant rightward moving of 4 pixels per frame.  All compared models are fine-tuned on the Sintel dataset. }
	\label{f7}
	\vspace{-0.4cm}
\end{figure}

Interestingly, we found that most of the existing models are unable robustly handle the samples with non-textured backgrounds contained in ChairSDHom.
To better illustrate, we produced similar samples of a rightward moving box with uniformly distributed texture to verify the generalization ability of the multiple models, and all models were taken from their official models fine-tuned on the Sintel dataset, including both unsupervised and supervised training. As shown in Fig. \ref{f7}, all compared models exhibit relatively poor results, whereas our ULDENet performs the best in comparison, which somehow demonstrates a behavior that is closest to the human interpretation of such scenarios.

\section{Conclusion}
\label{s7}
In this work, we explore the possibility of unsupervised learning-based optical flow estimation exposure to prolonged long-temporal dynamic environments and propose multiple unsupervised learning strategies based on multi-frame dynamic training. The contributions of this work are manifold: firstly, by partially following the well-known predictive coding mechanism, we construct a lightweight model with a spatial-temporal dual recurrent structure, which is demonstrated effectiveness in optical flow estimation in ablation studies; further, we design three temporal dynamic training enhancers, namely DOE, CVE, and SVE. The reasonable combination of the three improves the performance by approximately 20\% without incurring any computational and memory overheads. Remarkably, in coping with the intractable occlusion problem, we propose the temporal smoothing regularization that can supply a reliable reference for the occlusion region based on the motion prior from adjacent frames. In addition, we present an efficient method based on the Markov process in DOE to simulate the dynamic occlusion process, which, combined with the well-designed mixed supervision strategy, reduces the EPE of the OCC region on the Sintel training set by 16\%.

Combining all the proposals, we reach the state-of-the-art results with a fairly lightweight model for both the test set and the training set of multiple standard benchmarks.

\bibliographystyle{IEEEtran}
\tiny
\bibliography{szt}


\end{document}